\documentclass[10pt,journal,compsoc]{IEEEtran}
\newif\ifpeerreview

\peerreviewfalse

\usepackage[nocompress]{cite}
\usepackage{url}
\usepackage{amsmath,amssymb,graphicx}
\usepackage{xcolor}

\usepackage{lipsum} 

\usepackage{hyperref}

\usepackage[capitalize]{cleveref}
\crefname{section}{Sec.}{Secs.}
\Crefname{section}{Section}{Sections}
\Crefname{table}{Table}{Tables}
\crefname{table}{Tab.}{Tabs.}
\usepackage[switch]{lineno}
\usepackage{booktabs} 
\usepackage{multirow}

\usepackage{makecell}
\usepackage{enumitem}

\newcommand{\paperID}{59}

\title{Single-Step Latent Diffusion for Underwater Image Restoration}

\author{
Jiayi Wu$^{1,*}$, 
Tianfu Wang$^{1,*}$,
Md Abu Bakr Siddique$^{2}$, 
Md Jahidul Islam$^{2}$, \\ 
Cornelia Fermuller$^{1}$, Yiannis Aloimonos$^{1}$, Christopher A. Metzler$^{1}$ \\

$^{1}$University of Maryland \ $^{2}$University of Florida
\thanks{ $^*$ Equal contribution\\
Corresponding author: Tianfu Wang \href{mailto:tianfuw@umd.edu}{tianfuw@umd.edu}}
}

\begin{document}

\IEEEtitleabstractindextext{%
\begin{abstract}
Underwater image restoration algorithms seek to restore the color, contrast, and appearance of a scene that is imaged underwater. 
They are a critical tool in applications ranging from marine ecology and aquaculture to underwater construction and archaeology. 
While existing pixel-domain diffusion-based image restoration approaches are effective at restoring simple scenes with limited depth variation, they are computationally intensive and often generate unrealistic artifacts when applied to scenes with complex geometry and significant depth variation. 
In this work we overcome these limitations by combining a novel network architecture (SLURPP) with an accurate synthetic data generation pipeline. 
SLURPP combines pretrained latent diffusion models---which encode strong priors on the geometry and depth of scenes---with an explicit scene decomposition---which allows one to model and account for the effects of light attenuation and backscattering. 
To train SLURPP we design a physics-based underwater image synthesis pipeline that applies varied and realistic underwater degradation effects to existing terrestrial image datasets.
This approach enables the generation of diverse training data with dense medium/degradation annotations. We evaluate our method extensively on both synthetic and real-world benchmarks and demonstrate state-of-the-art performance. Notably, SLURPP is over $200\times$ faster than existing diffusion-based methods while offering  $\sim 3 dB$ improvement in PSNR on synthetic benchmarks. It also offers compelling qualitative improvements on real-world data. 
Project website \href{https://tianfwang.github.io/slurpp/}{https://tianfwang.github.io/slurpp/}.
\end{abstract}

\begin{IEEEkeywords} 
Computational Imaging, Underwater Restoration, Denoising Diffusion, Foundational Models
\end{IEEEkeywords}
}

\maketitle

\IEEEraisesectionheading{
  \section{Introduction}\label{sec:introduction}
}
\IEEEPARstart{U}{nderwater} Image Restoration is a critical task due to the widespread degradation of visual quality in submerged environments caused by light absorption, scattering, and color distortion. These degradations significantly hinder visual perception, making it diffcult for computer vision systems to interpret underwater scenes accurately. Restoring underwater images is essential for a variety of applications, including marine biology research, underwater archaeology, environmental monitoring, autonomous underwater vehicle (AUV) navigation, and underwater robotics. However, underwater image restoration is inherently difficult due to the complex optical properties of water~\cite{akkaynak2018revised}, which differ with depth, turbidity, and lighting conditions. Traditional model-based methods rely on physical priors~\cite{siddique2024aquafuse}, but often struggle with generalization across diverse underwater scenes. Recently, learning-based approaches have shown promising results by leveraging data-driven priors, yet they still face challenges such as a lack of ground truth data, domain shift, and poor interpretability. These limitations highlight the need for more robust and generalizable methods that can adapt to complex degradations of underwater scenes.
\begin{figure*}[t]
  \centering
   \includegraphics[width=\linewidth]{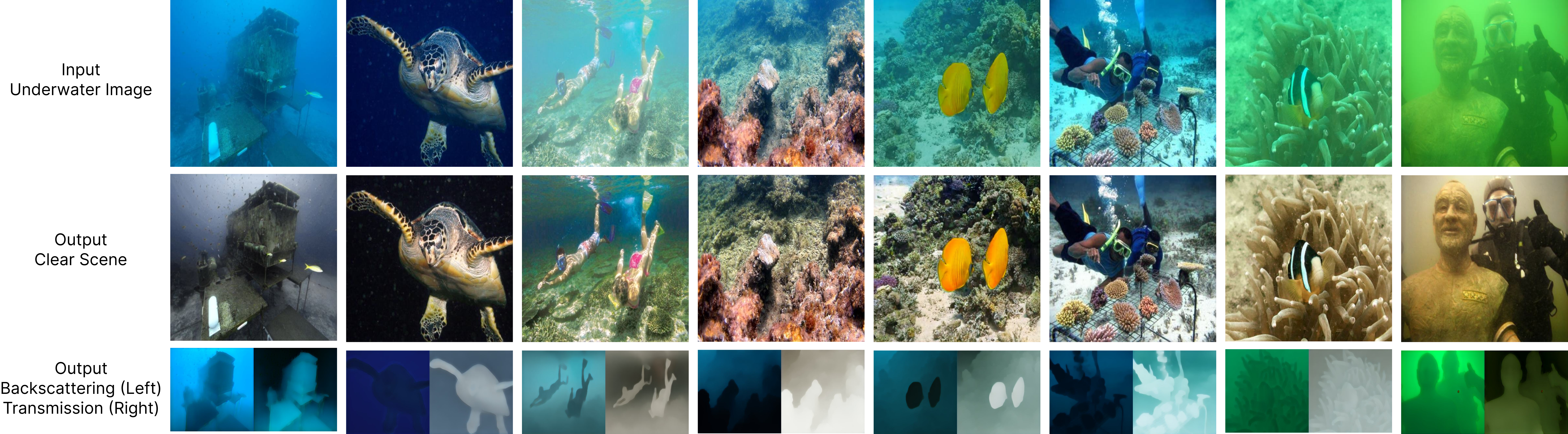}
   \vspace{-20pt}
   \caption{\textbf{Real-world underwater restoration using our method.} We develop a single-step underwater restoration method that leverages pretrained latent diffusion priors. Given an underwater input image (top row), our method jointly predicts the clear image (middle row), and the per-pixel underwater medium parameters, specifically the backscattering (bottom row left) and transmission (bottom row right) parameters. In this figure, we present real-world results using images from the UIEB~\cite{li2019UIEB} underwater dataset. We show that our method can robustly restore underwater images in a variety of different scenes and water conditions.   }

   \label{fig:1_teaser}
\end{figure*}

Modern text-to-image latent diffusion models~\cite{rombach2021latentdiffusion, ho2020denoising_ddpm}, trained on massive online datasets~\cite{schuhmann2022clip}, have demonstrated remarkable generative capabilities. Crucially, the rich knowledge encoded within extends significantly beyond image synthesis. Recent works have highlighted this by successfully repurposing pretrained latent diffusion models for challenging computer vision tasks, including dense prediction problems such as monocular depth estimation and image intrinsic decomposition~\cite{ke2023marigold, fu2024geowizard,wang2024flash,chen2025uni}. Such successes strongly suggest that these models implicitly capture a sophisticated understanding of scene geometry and intrinsic properties~\cite{wu2024viewactive}, learned inherently from the structure present in their massive training corpora~\cite{xiongevent3dgs}.

To this end, underwater image restoration presents a unique challenge, aiming to recover clear visual scene appearances distorted by wavelength-dependent scattering and absorption effects inherent to the water medium. 
Fundamentally, it requires a joint estimation of both the clear image and the physical medium parameters governing underwater visual degradation.
Insights from underwater imaging~\cite{akkaynak2018revised,akkaynak2019sea} suggest that these two components—scene content and water medium—can serve as mutual cues.
We propose that the rich generative priors encoded in pretrained diffusion models offer a powerful framework to address both aspects of this problem. Specifically, the target clear images often depict natural scenes, aligning well with the distribution of content that such models are trained on. 
Moreover, backscattering and attenuation effects exhibit strong correlations with scene depth. Notably, recent advances in monocular depth estimation using pretrained latent diffusion~\cite{yu2022udepth,ke2023marigold,fu2024geowizard,martingarcia2024diffusione2eft} reveal that these models inherently capture robust depth priors. This suggests a promising opportunity for modeling depth-dependent water medium parameters, offering a unified approach to underwater image restoration that is both data-efficient and physically grounded.

Building on this insight, we introduce \textbf{SLURPP}: \textbf{S}ingle-step \textbf{L}atent \textbf{U}nderwater \textbf{R}estoration with \textbf{P}retrained \textbf{P}riors. 
Our method is a {latent-diffusion-based restoration framework} that offers a simple yet effective \textbf{single-step solution} for underwater image restoration. 
SLURPP is simple in that it performs direct, physically informed fine-tuning of the underlying latent diffusion model for the task of single-step underwater image restoration, without explicit auxiliary task training, such as dedicated depth prediction.
We design a dual-branch architecture to jointly estimate the clear scene and the dense depth-dependent water medium.
Crucially, we inject distinct diffusion priors into each branch, tailored to their respective tasks. 
Our method enables robust and data-efficient restoration across a wide range of underwater conditions (\cref{fig:1_teaser}), overcoming challenges posed by the scarcity of real-world underwater datasets 
and the difficulty of obtaining paired ground truth data. 

\vspace{1mm}
\noindent
Our main contributions are summarized as follows:
\begin{enumerate}[label={\arabic*)},nolistsep,leftmargin=*]
    \item We propose a novel underwater image restoration approach that leverages the foundational visual and geometric priors embedded in pretrained latent diffusion models.
    Our method jointly estimates the clear scene and water medium properties \textit{in a single step}. By direct fine-tuning a dual-branch architecture tailored for disentangling image content from depth-dependent waterbody effects, our SLURPP method achieves efficient and high-quality restoration across diverse underwater scenes. 
    \item We develop a physically grounded and computationally efficient underwater image simulation pipeline, built upon the standard underwater image formation model and informed by real-world optical measurements of underwater environments. This pipeline enables the synthesis of high-quality, realistic paired training data by simulating diverse underwater conditions—including varying water types, depths, and lighting—on top of large-scale, easily accessible terrestrial image datasets.

    \item By fine-tuning on our simulated dataset, our method effectively adapts the strong generative priors of pretrained latent diffusion models to the specific task of underwater image restoration. In contrast to prior approaches that rely on pixel-space diffusion, our framework operates in a more compact and expressive latent space, enabling fast single-step inference that is over 200$\times$ faster than previous diffusion methods, while also supporting the restoration of higher-resolution images with greater visual fidelity. This efficiency, coupled with improved restoration quality, highlights the practical and technical benefits of leveraging latent generative priors for real-world underwater imaging applications.
    
\end{enumerate}

\section{Related Work}
\subsection{Underwater Image Restoration and Enhancement}
Underwater image restoration and enhancement, although closely related, address different aspects of image quality improvement.
Enhancement methods improve visual quality by adjusting contrast, color, and brightness without modeling physics~\cite{islam2020fast}, using techniques such as histogram equalization, Retinex, and local contrast adjustments~\cite{islam2020sesr}. While computationally efficient, these enhancement methods often produce visually appealing but physically implausible results.
In contrast, restoration methods aim to recover true scene radiance by modeling underwater light propagation~\cite{wu2024marvis}, accounting for absorption, scattering, and wavelength-dependent attenuation~\cite{akkaynak2019sea}.

Traditional restoration methods rely on handcrafted priors and physical models to estimate and mitigate degradations~\cite{siddique2024aquafuse,li2019UIEB,wu20233d,wu2023low}. Deep learning has significantly advanced underwater image restoration and enhancement by offering adaptive solutions to learn complex mappings from data. 
\cite{li2019UIEB} established the first CNN-based benchmark. 
\cite{fabbri2018UGAN} and~\cite{islam2020funiegan} applied adversarial training for color and detail enhancement. 
\cite{huang2023semiUIR} leverages unlabeled data 
 pseudo-labeling and contrastive learning.
Transformer approaches such as~\cite{peng2024histoformer} and~\cite{khan2024phaseformer} introduced histogram and phase-based self-attention mechanisms.~\cite{sharma2021deepwavenet} uses wavelength-aware networks that enhance restoration using adaptive receptive fields and attentive skip connections. 
Most relevant to our work,~\cite{nathan2024osmosis} integrates RGBD diffusion priors with a physically-based sampling scheme.

\subsection{Diffusion Models for Computer Vision Tasks}
Recent large latent diffusion models (LDMs)~\cite{rombach2021latentdiffusion}, trained on massive online datasets of text-image pairs~\cite{schuhmann2022clip}, can generate diverse and photorealistic images with a text prompt, inspiring large attention in the computer vision community.
The first extensions on LDMs focus on controllable image generation using additional conditions such as depth, inpainting, and segmentation maps~\cite{ zhang2023controlnet, meng2021sdedit, wang2023breathing, jia2024dginstyle, wang2024consistency, cai2025parametric}. 
Since then, several works have repurposed LDMs for non-generative tasks, such as monocular depth and normal estimation~\cite{ke2023marigold, fu2024geowizard,ye2024stablenormal, martingarcia2024diffusione2eft} and image intrinsic decomposition~\cite{chen2025uni}.
LDMs have also been used in the image restoration tasks such as deblurring~\cite{whang2022deblurring}, super-resolution~\cite{mei2025power}, and flash removal~\cite{wang2024flash}. 
On underwater image restoration, we note that Osmosis~\cite{nathan2024osmosis} is the first work to leverage diffusion priors. However, they only use a pixel space RGBD diffusion model trained from ImageNet~\cite{deng2009imagenet, dhariwal2021diffusion} with limited generation resolution and much less scale and generative capacity compared to current pretrained latent diffusion models~\cite{rombach2021latentdiffusion}. 
Additionally,~\cite{nathan2024osmosis} uses a DDPM~\cite{ho2020denoising_ddpm} sampling scheme that requires 1000 inference steps, while our method enables single-step inference.

\vspace{-15pt}
\section{Proposed Method}
\subsection{Preliminaries}

\subsubsection{Underwater Image Formation}
The Jaffe-McGlamery (JM) model~\cite{jaffe1990JM} serves as a fundamental framework in underwater imaging, providing a mathematical representation of the complex processes of light absorption and scattering in aquatic environments, which significantly influence the visual appearance of submerged objects. Subsequently,~\cite{akkaynak2018revised} introduced a revised underwater image formation model that incorporates variations in attenuation coefficients between direct transmission and backscatter to enhance the accuracy of underwater image correction techniques. A widely adopted formulation of the underwater image formation process is expressed by the following equation:
\begin{equation}
\label{eq:UIFM_Eq}
I_c = J_c \cdot e^{-\beta^D_c z} + B_c^\infty \cdot (1 - e^{-\beta^B_c z}),
\end{equation}
where \( c \) $\in$ \{R, G, B\} represents the color channel; \( I \) represents the image captured underwater by the camera of a scene at distance \( z \); \( J \) denotes the clear scene that would have been captured in the absence of water along the line of sight; and \( B^\infty \) refers to the water color at infinity, commonly referred to as the background light. The two parameters \( \beta^D \) and \( \beta^B \) represent the attenuation and backscatter coefficients, respectively.
\begin{figure}[t]
  \centering
   \includegraphics[width=0.99\columnwidth]{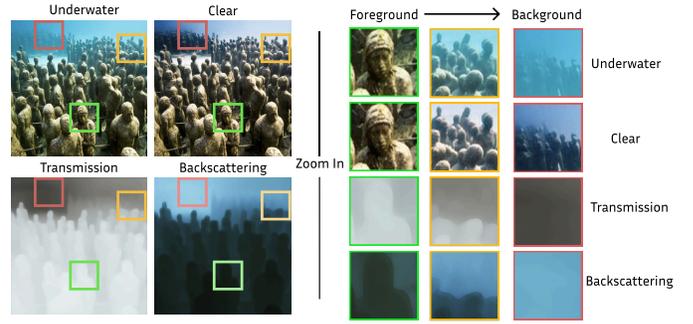}

   \caption{\textbf{Our method captures the depth-varying change of water medium properties.} In this figure we demonstrate the depth-dependent nature of the underwater medium effects and show that our method can correctly capture this in our medium predictions. We can see in the zoomed-in regions of the input underwater image (right first row), that the water medium effects increase as we move from the foreground to the background. This illustrates that the water medium effects are strongly correlated with scene depth. Our model can recover both the clear image (right second row) while capturing the depth-correlated transmission and backscattering effects (right bottom rows). Our model predicts as the scene depth increases, the backscattering becomes stronger while the transmission becomes weaker. This prediction aligns with the observed medium phenomenon in the input underwater image. }

   \label{fig:depth_variance}
\end{figure}

\begin{figure*}[t]
  \centering
   \includegraphics[width=0.8\linewidth]{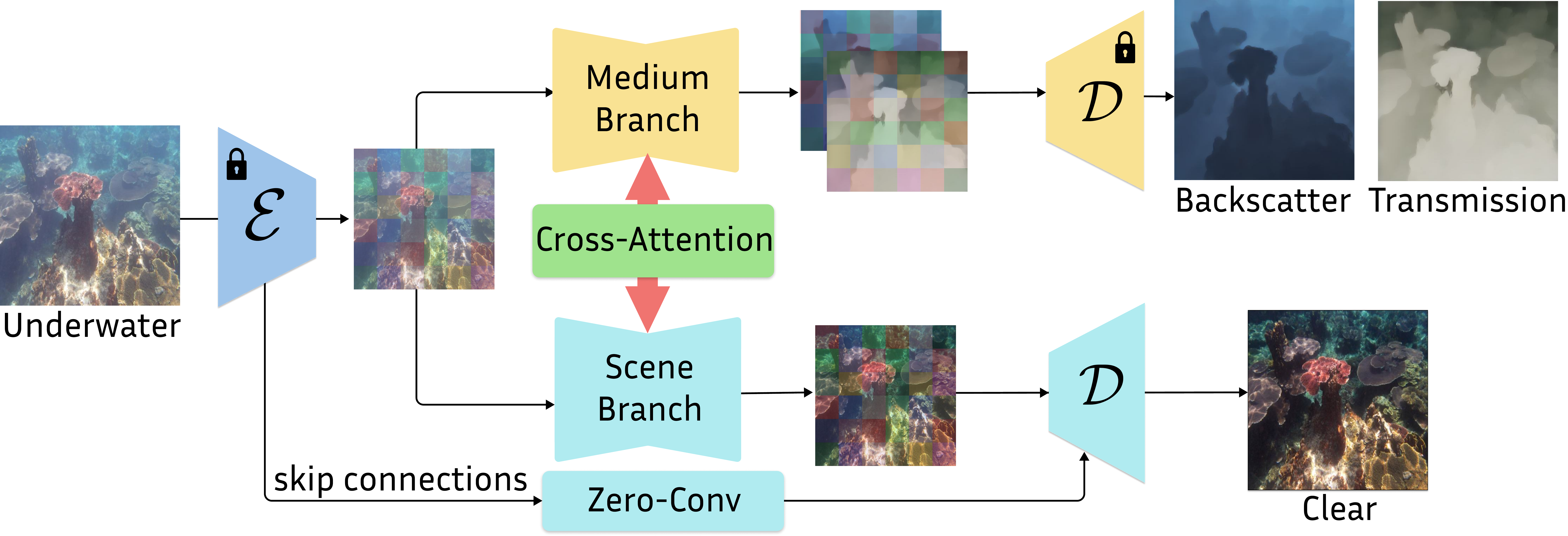}
   \caption{\textbf{Pipeline overview of our single-step dual-branch underwater restoration method.} Our pipeline takes in an underwater image and aims to predict a clear image without water effects, along with the transmission and backscattering properties of the water medium in a single step. The input image is first encoded into latent space using the frozen VAE from pretrained Stable Diffusion (SD)~\cite{rombach2021latentdiffusion}. This latent image is then fed into two UNet~\cite{unet} branches: the scene branch to predict the clear scene, and the medium branch to predict the wavelength and depth-dependent medium effects. The two branches use different pretrained diffusion priors~\cite{rombach2021latentdiffusion, ke2023marigold} that fit their respective prediction modalities while exchanging mutual cues through a cross-attention mechanism. The UNets then predict the scene and medium latent images {in a single step}. To output the predictions, the attenuation and backscattering latent images are decoded using the standard SD decoder, while the clear image is decoded with a cross-latent decoder fine-tuned by incorporating high-frequency details passed from the input image through skip connection layers of the encoder. }

   \label{fig:inference}
\end{figure*}
\vspace{-5pt}

\subsubsection{Latent Diffusion Model and Diffusion Fine-Tuning}
Denoising Diffusion Probabilistic Models 
 (DDPMs)~\cite{song2020denoising_ddim, song2020score} are generative models that learn data distributions by reversing a Markovian process forward process, in which data is gradually corrupted by Gaussian noise over several steps.
Early works on diffusion-based image generation are directly trained on RGB pixel space~\cite{song2020denoising_ddim, dhariwal2021diffusion}, which imposes large computational and memory requirements for training and inference. 
Latent Diffusion Models (LDMs) such as Stable Diffusion (SD)~\cite{rombach2021latentdiffusion} shift the diffusion process in a low dimensional latent space defined by a variational autoencoder (VAE)~\cite{Kingma2013VAE}.
The VAE improves computational efficiency by contracting the image's spatial dimension, while expanding the feature dimension helps encode high-level features and creates a smoother sampling landscape for effective generation.
The computational and modeling advantages of the LDMs lead to their wide adoption in image generation. 
However, LDMs are still slow due to their need for iterative denoising during inference, with work on few-step sampling~\cite{song2020denoising_ddim, luo2023latent, song2023consistency, salimansprogressive} trading inference time with generation quality.

Recent work repurpose LDMs for computer vision tasks~\cite{ke2023marigold,wang2024flash, fu2024geowizard}, achieving impressive results.
However, these works still model the estimation process as conditional generation based on additional image input. 
As such, they still need iterative denoising during inference. 
Inspired by recent theoretical and empirical progress in single-step diffusion~\cite{karras2022elucidating, ye2024stablenormal, martingarcia2024diffusione2eft}, we hypothesize that the iterative denoising formulation is less crucial for underwater image restoration, where the distribution of the predicted restored image is narrow and peaks at the ground truth, compared to text-to-image generation, where there is a wide distribution of plausible images for a text prompt.
As such, we choose a pipeline that removes stochasticity from the training and inference process and directly predicts restored properties in a single step.
We show in our experiment that this single-step formulation does not degrade performance and could even outperform iterative denoising versions of our method, due to the additional advantage of the ability to directly supervise in the image space for single-step training. 

\vspace{-10pt}
\subsection{Problem Setting}

From the underwater image formation model \cref{eq:UIFM_Eq}, two key insights emerge: first, 
underwater light attenuation and scattering exhibit a strong dependence on both wavelength and propagation distance; second, the occluding backscatter layer, which degrades image clarity, is inherently independent of the scene content. 
Based on these insights, we formulate our restoration task as the joint estimation of both the restored clear image $J$, and medium-related parameters, including transmission $T$ and backscattering $B$, of the input underwater image $I$.
The output of our prediction should fit the underwater image formation model specified in \cref{eq:UIFM_Eq}, where we now write as our dense scene-medium decomposition formulation:
\begin{equation}
\label{eq:formulation}
    I_c = J_c \cdot T_c + B_c
\end{equation}

Compared to the imaging model of \cref{eq:UIFM_Eq}, we see that for each channel $c$, we have the relation $T_c = e^{-\beta^D_c z}$ and $B_c =  B_c^\infty \cdot (1 - e^{-\beta^B_c z})$, showing that both medium predictions are highly correlated with the scene depth. We represent medium properties using two three-channel images, \( T \) and \( B \), rather than separately estimating depth \( z \) and water parameters \( \beta^D \), \( \beta^B \), and \( B^\infty \) as in previous methods, for three main reasons.  

First, to incorporate diffusion priors, our pipeline needs to preserve the architecture of pretrained latent diffusion models, which are naturally suited for high-dimensional dense signals such as images.  Second, previous methods often assume medium homogeneity (i.e., a spatially uniform \( \beta \)) and even reduce the number of unknown parameters by setting \( \beta^D = \beta^B \). Our approach avoids these simplifying assumptions, enabling more robust estimations. Finally, dense medium parameters are depth-related but represented as bounded image intensities, unlike raw depth values with infinite range. This makes them more robust for reconstructing scenes with large depth variations, as medium images are more robust to depth estimation errors in the distant background. We illustrate the effectiveness of our formulation in~\cref{fig:depth_variance}.

\subsection{Our Core Idea}
\vspace{-2pt}

The core idea of our method can be broken down into three key points: foundational model prior, single-step task specific fine-tuning, and physically-accurate training data. 
Effective underwater restoration requires capturing two aspects: a clear image that resembles natural scenes and the water medium properties correlated with scene depth. 
With this in mind, we leverage current pretrained latent diffusion models to provide foundational natural image priors for clear image prediction and depth priors for medium prediction.
We design a dual-branch architecture (\cref{fig:inference}) for joint scene and medium prediction, consisting of a scene branch for content restoration and a medium branch for estimating pixel-level medium parameters.
The scene branch is initialized from a pretrained text-to-image diffusion model~\cite{rombach2021latentdiffusion} containing strong priors on natural images,
while the medium branch is initialized from a pretrained affine-invariant monocular depth diffusion model~\cite{ke2023marigold}.
At the training level, our framework and fine-tuning strategy are designed to incorporate the prior knowledge of the underwater image formation model while enabling fast, high-quality single-step inference. We introduce inter-branch cross-attention to exploit the complementary relationship between the clear image and water medium, allowing them to serve as mutual cues during prediction. Additionally, our training objective includes a reconstruction loss that explicitly encourages the outputs to adhere to the dense scene–medium decomposition described in~\cref{eq:formulation}.
At the data level, we train our model using physically accurate data, enabling it to learn the underwater image formation process for robust predictions. 
Addressing the lack of large-scale real paired underwater datasets, we synthesize training data by applying a physically accurate formation model to diverse terrestrial images as clean sources. 
Our physically accurate underwater image synthesis pipeline uses a carefully optimized medium-related parameter generation strategy to synthesize high-quality training data.

\vspace{-5pt}

\subsection{Physics-based Diverse Underwater Data Synthesis}
\label{sec:data}
\begin{figure}[t]
  \centering
   \includegraphics[width=0.9 \columnwidth]{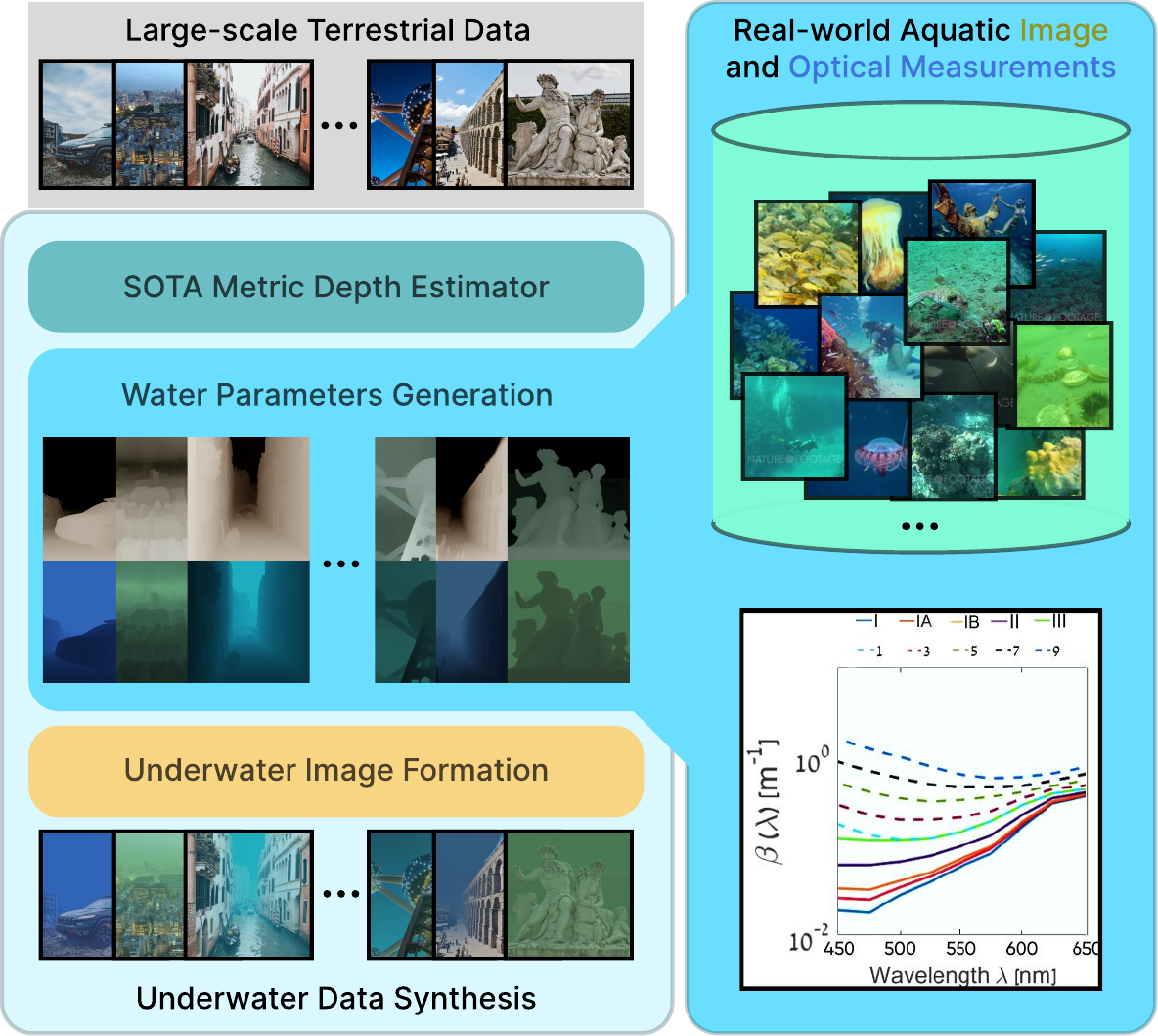}
   \vspace{-2pt}
   \caption{\textbf{Physically accurate underwater image synthetics pipeline for diverse data generation.} 
    Our model is trained on realistic underwater images synthesized from large-scale terrestrial data using precise modeling of depth, attenuation, and background light for physically accurate results.
   We generate accurate metric depth maps using a state-of-the-art metric depth estimator~\cite{bochkovskii2024depth}.
   We sample attenuation values based on real-world water measurements~\cite{Solonenko:15,akkaynak2018revised} (bottom-right, reproduced from~\cite{akkaynak2018revised}).
    We source diverse, realistic background light estimated from real-world underwater images~\cite{ULAP} (top-right).
   These generation strategies enhance the realism and quality of our synthetic training data. 
   }

   \label{fig:data_gen}
\end{figure}
\begin{figure}[t]
  \centering
   \includegraphics[width=0.97\columnwidth]{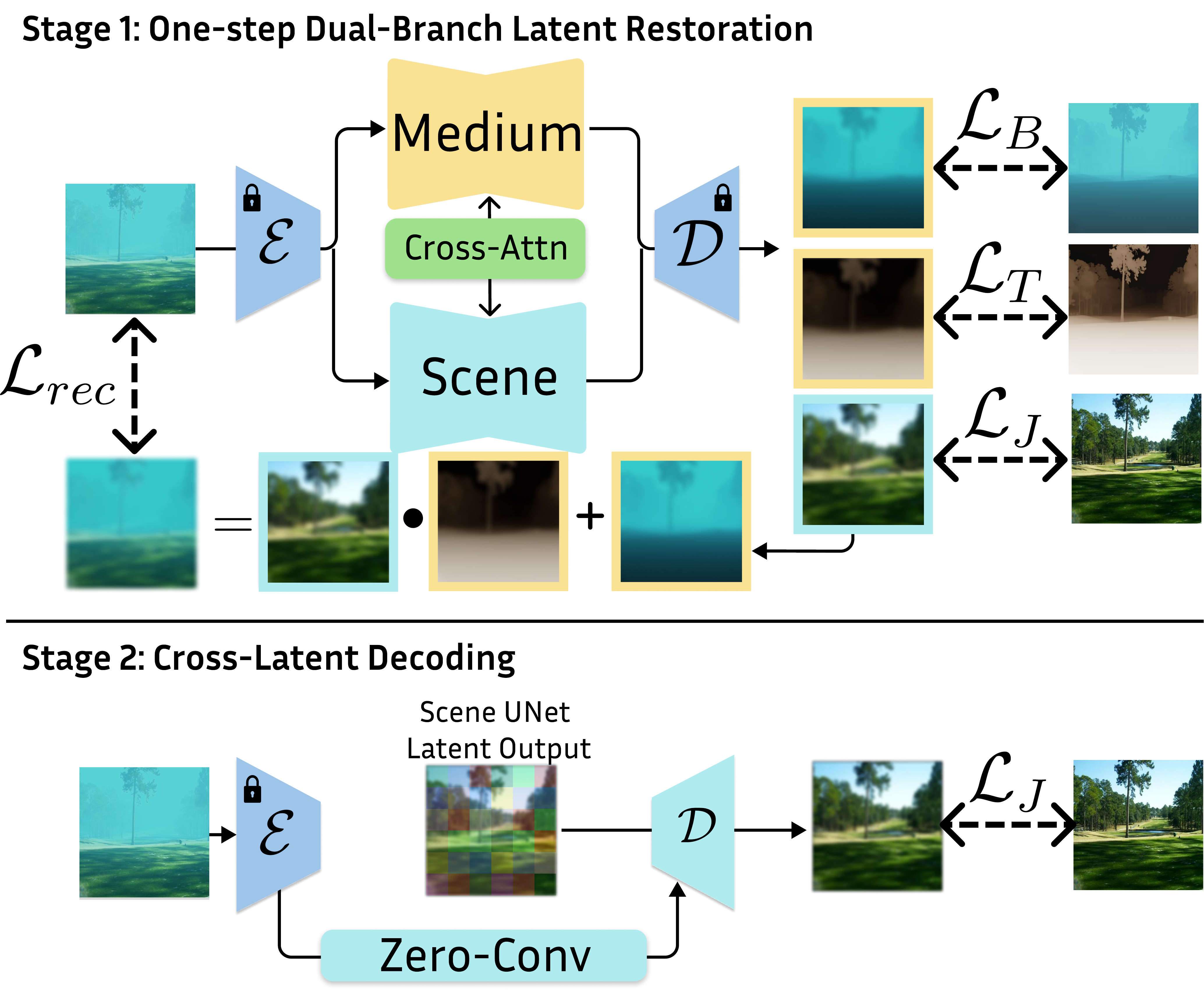}
   \caption{\textbf{Two-stage training procedure of our method.} In our first stage, we train our dual-branch UNets with inter-branch cross-attention to directly predict the latent images of the clear scene $J$, as well as medium transmission $T$ and backscattering $B$. The latent outputs are decoded and supervised with their respective ground truths using image losses. We also use the reconstruction loss to guide the predicted outputs to respect the underwater image formation model. For stage 2 cross-latent decoding, we fine-tune the decoder and additional zero convolution skip connections to transfer high-frequency details from the input underwater image to the restored image. }

   \label{fig:training}
\end{figure}
Real-world underwater datasets are scarce and typically lack ground truth, hindering the development of generalizable image restoration models. 
3D simulators, while an alternative, often suffer from high modeling costs, limited scene diversity, and a large domain gap compared to reality. 
Consequently, synthesizing physically plausible underwater images by applying imaging models to large-scale terrestrial data has become a standard approach in the field.

The underwater image formation model 
 \cref{eq:UIFM_Eq} shows that the degradation caused by the scattering medium is mainly governed by the depth \(z\), the attenuation coefficient \(\beta\), and the background light \(B^\infty\).  Accurately and diversely generating these parameters is crucial for realistic synthetic data. 
 Unlike prior methods that approximate the formation model, we optimize each parameter to achieve fine-grained rendering of light scattering and attenuation (\cref{fig:data_gen}). 

Obtaining the depth $z$ is challenging as the underwater image formation model \cref{eq:UIFM_Eq} requires the absolute depth in meters. While RGBD datasets~\cite{geiger2013kitti} provide metric depth, they are often sparse, lack scene diversity, and are costly to acquire. As a result, prior works~\cite{nathan2024osmosis,ULAP} often rely on monocular depth predictions with manual normalization. However, normalization without camera intrinsics introduces scale errors, limiting the reliability of depth and downstream water medium generation. To address this, we leverage Depth Pro \cite{bochkovskii2024depth}, a recent advancement in metric monocular depth that directly predicts focal length from the input image, enabling accurate metric depth estimation. This allows us to generate dense, reliable, and diverse per-pixel metric depth maps from large-scale image collections.

While the values of the attenuation coefficient \( \beta \) and background light \(B^\infty\) are theoretically arbitrary, they are intrinsically constrained on water type. 
To ensure consistency with real-world aquatic environments, our generation strategy is informed by global-scale underwater optical measurements and extensive real-world data priors. 
Specifically, we first randomly sample from the 10 Jerlov's water types~\cite{Solonenko:15}, and then using the sample's measured coefficients at 600nm, 525nm, and 475nm to guide the RGB intensities  of \( \beta \) respectively. 
To avoid unrealistic over or under degradation of the image, we bound the attenuation \( \beta \), and resample when excessive information loss occurs in the generated image.

We source diverse, realistic background light \(B^\infty\) values by extracting them from real underwater images using ULAP~\cite{ULAP}, where we swap its monocular depth component with DepthPro~\cite{bochkovskii2024depth} for more precise  background light estimation. 
The extracted background lights are then clustered (K=10 K-means in the Lab color space 'ab' channels) into perceptually distinct subsets representing different water types. 
During synthesis, we first randomly select a light cluster and then randomly sample within it to obtain the background light \(B^\infty\).

Using our generation strategies for depth \( z \), attenuation \( \beta \), and background light \( B^\infty \), and applying the underwater image formation model  \cref{eq:UIFM_Eq}, we can efficiently generate high-quality paired data required to train our model.

\begin{figure*}[t]
  \centering
   \includegraphics[width=0.99\linewidth]{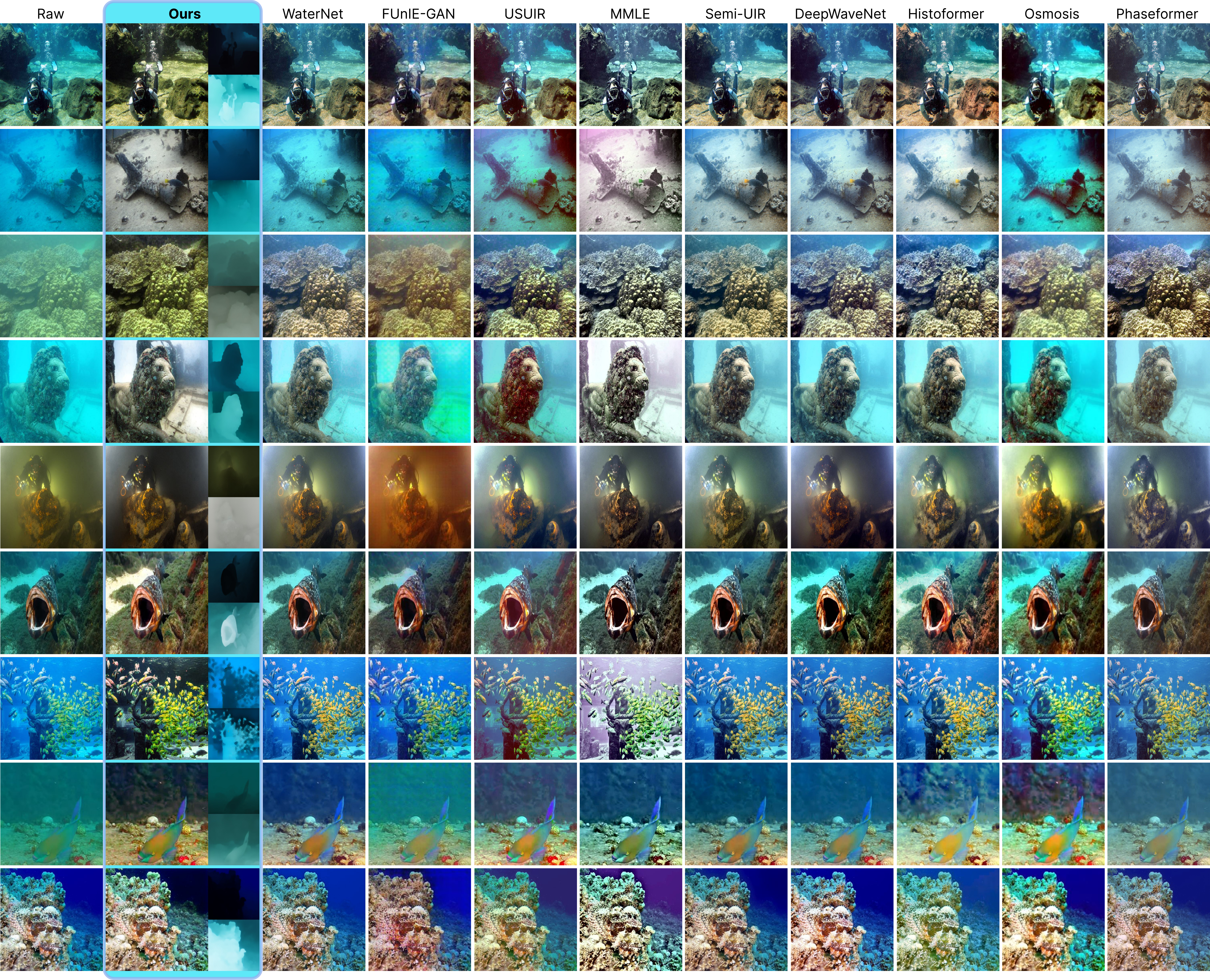}
   \caption{\textbf{Qualitative comparisons of restoration results in USOD10K~\cite{hong2023usod10k} UIEB~\cite{li2019UIEB}}. We show extensive comparisons against previous methods~\cite{li2019UIEB, islam2020funiegan, fu2022USUIR, zhang2022MMLE, huang2023semiUIR, sharma2021deepwavenet, peng2024histoformer, nathan2024osmosis, khan2024phaseformer}. 
   As illustrated in the comparison, previous methods often struggle to achieve physically consistent restoration across the entire scene, and may even exhibit unnatural changes in water body color. In contrast, our method (second column to the left) achieves physically-consistent restoration across scenes of varying depth, with notably improved performance in severely degraded distant areas.
   Furthermore, our method accurately estimates per-pixel medium parameters and enables precise and faithful scene restoration across diverse water types and color profiles.
   }

   \label{fig:qual_comp}
\end{figure*}
\subsection{Pipeline Architecture}
Our proposed pipeline (\cref{fig:inference}) addresses underwater image restoration by jointly predicting the clear scene image, free from water effects, along with the transmission and backscattering properties characterizing the water medium. Initially, the input underwater image undergoes encoding into the latent space via the frozen pretrained Stable Diffusion (SD) VAE encoder~\cite{rombach2021latentdiffusion, Kingma2013VAE}. This latent representation serves as input to a dual-branch architecture comprising two UNets~\cite{unet} connected with inter-branch cross-attention: a scene branch tasked with predicting the clear scene latent, and a medium branch predicting latent images of depth-dependent attenuation and backscattering. 
Both branches predict their respective latent images in a single step. Finally, the decoding process differs based on the output type: the attenuation and backscattering latent variables are decoded using the standard SD decoder. In contrast, the clear scene latent is decoded using the cross-latent decoder, which is fine-tuned to incorporate high-frequency details passed from the original input image via skip connections originating from the VAE encoder.

We now describe the training process illustrated in~\cref{fig:training}. We train the single-step restoration and cross-latent decoder in two stages.
We note that during inference the frozen encoder can be trivially modified to introduce cross-latent decoding for unified single-step inference.

\subsubsection{Stage 1: Single-step Restoration Fine-Tuning}
Typical conditional diffusion fine-tuning~\cite{ke2023marigold, wang2024flash} first converts both the input and the output ground truth images to latent space and injects the output latent image with a random proportion of Gaussian noise. The latent diffusion UNet is then fine-tuned to predict the injected noise, given the input latent and noisy output latent as inputs. 
This training recipe is tailored for iterative denoising inference, but performs poorly for few-step inference. Additionally, the loss is applied to the noisy latent image that is uninterpretable and cannot leverage structural and perceptual image supervision.
In our single-step fine-tuning, the diffusion UNet simply takes in the input latent concatenated with the zero image, which is the mean of the pure Gaussian noise distribution, and learns to directly predict the output latent image in one pass. 
The output latent image can then be decoded into RGB space where we can compare with the ground truth output using the pixel-space image loss:

\begin{equation}
\label{eq:rgb_loss}
\mathcal{L}= \mathcal{L}_1 + \mathcal{L}_{SSIM} + \mathcal{L}_{LPIPS}.
\end{equation}

We use this training strategy to jointly train the two UNet branches (with cross-attention), with the encoder and decoder both frozen to their pretrained weights. We apply the image loss in \cref{eq:rgb_loss} to all output modalities compared to their respective ground truth.
Additionally, we combine the predicted images using the  dense scene-medium decomposition formulation in \cref{eq:formulation}, and the apply the image loss in \cref{eq:rgb_loss} to the input image as a self-supervised reconstruction loss $\mathcal{L}_{UIFM}$. Our final loss can be written as 

\begin{equation}
\label{eq:total_loss}
\mathcal{L}_{total} = \lambda_{J}\mathcal{L}_{J}+ \lambda_{T}\mathcal{L}_{T} + \lambda_{B}\mathcal{L}_{B} + \lambda_{L}\mathcal{L}_{UIFM}.
\end{equation}
We use $\lambda_{J} = 1, \lambda_{T} = \lambda_{B} = 0.5, \lambda_{L} = 0.4$.

We note that \( L_{UIFM} \) plays a key role in mitigating the domain gap introduced by synthetic training data. 
While our improved data generation pipeline approximates real-world degradation, it remains limited by the lack of dense optical measurements in real underwater environments, leading to simplifications in light propagation and attenuation modeling. These assumptions introduce discrepancies between synthetic and real images, such as non-uniform media or mismatched scattering and attenuation coefficients. 
By guiding the model to learn from intrinsic data consistency rather than reply solely on synthetic labels, the reconstruciton loss $L_{UIFM}$ improves robustness against synthetic data biases and enhances generalization to diverse real-world underwater environments.

\begin{figure}[t]
  \centering
   \includegraphics[width=\columnwidth]{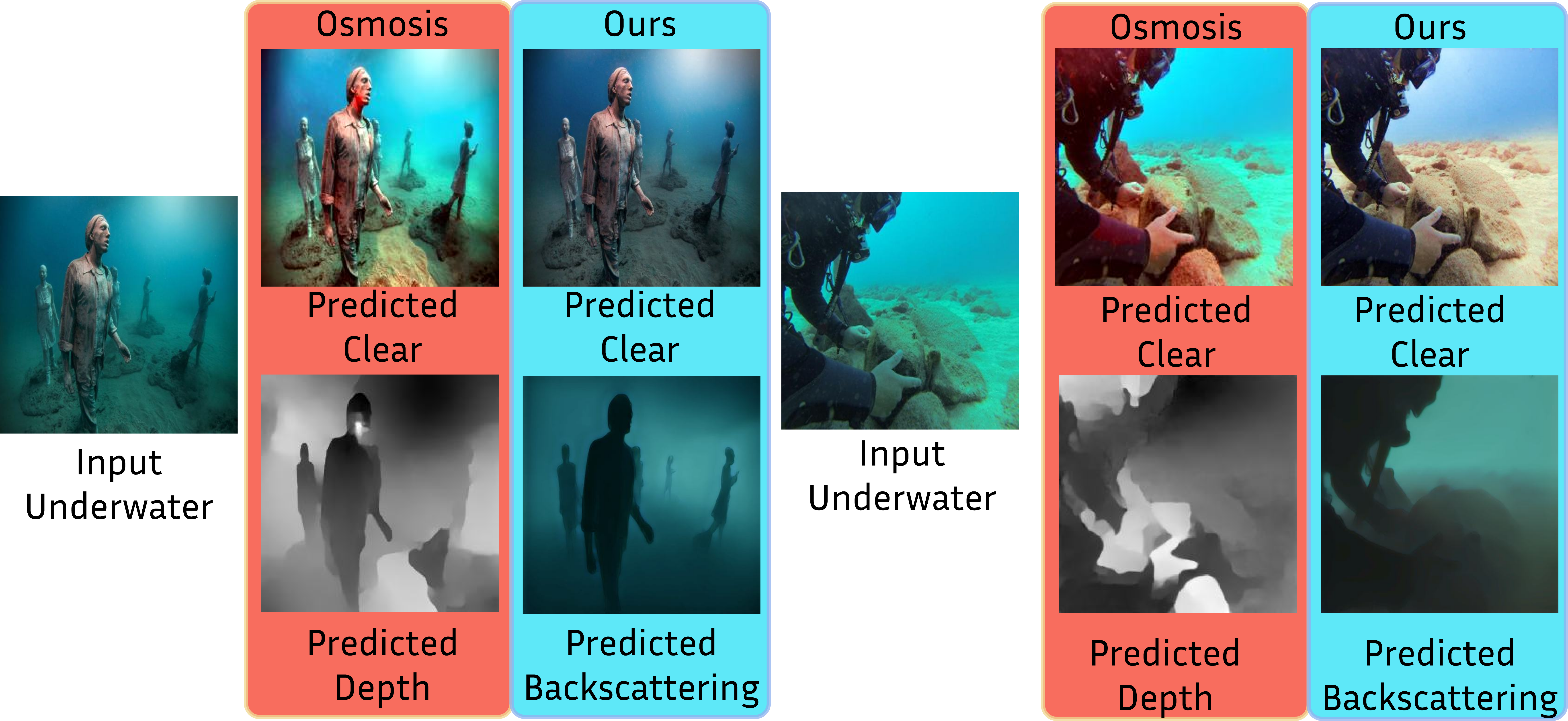}
   \caption{\textbf{Comparison with Osmosis~\cite{nathan2024osmosis} on the UIEB dataset~\cite{li2019UIEB}.} In this figure we show the predicted clear image and medium-related parameters for our method and Osmosis. In the medium visualization of both methods, objects in the foreground have lower depth/backscattering, while background objects have higher depth/backscattering. Osmosis highly depends on accurate depth estimation, incorrect depth (such as the diver's face region in the right image) leads to unrealistic restoration with spurious color artifacts. Our scene-medium separation formulation leverages depth priors indirectly through water medium prediction, and we obtain much better quality predictions for both clear restoration and depth-dependent medium parameters.}

   \label{fig:osmosis_cmop}
\end{figure}

\subsubsection{Stage 2: High-frequency Preservation Decoding}
Single-step latent restoration can already effectively restore the clean image.
However, due to limitations of the vanilla SD decoder~\cite{rombach2021latentdiffusion}, we still observe blurriness and hallucinations in high-frequency details such as text. Following previous diffusion-based restoration methods~\cite{wang2024flash}, we use cross-latent decoding with additional zero convolution skip-connections to transfer high-frequency details from the underwater input to the clear image. Once the dual-branch diffusion is trained, we use pairs of underwater image and the latent image output of the scene branch UNet for the second stage corss-latent deocder training, where we fine-tune only the zero convolution and the decoder. 
This training is supervised using the same image loss in \cref{eq:rgb_loss} between the decoded image and the ground truth image.

\section{Experimental Results}
\begin{table}[t]
\centering
\caption{\textbf{Quantitative comparison on USOD10K~\cite{hong2023usod10k} and UIEB~\cite{li2019UIEB} datasets using UIQM~\cite{panetta2015UIQM} and MUSIQ~\cite{ke2021musiq} reference-free metrics.} Due to the lack of ground truth clear images for real-world underwater datasets, we use reference-free metrics that measure the clear image quality as an assessment to restoration effectiveness. For both datasets our method achieves the best reference-free metric performance. }
\renewcommand{\arraystretch}{1.2}
\setlength{\tabcolsep}{8pt}
\resizebox{\linewidth}{!}{\begin{tabular}{lcccc}
\toprule
\multirow{2}{*}{\textbf{Method}} & \multicolumn{2}{c}{\textbf{USOD10K~\cite{hong2023usod10k}}} & \multicolumn{2}{c}{\textbf{UIEB~\cite{li2019UIEB}}} \\
\cmidrule(lr){2-3} \cmidrule(lr){4-5}
 & UIQM$\uparrow$ & MUSIQ$\uparrow$ & UIQM$\uparrow$ & MUSIQ$\uparrow$ \\
\midrule
WaterNet (TIP 2019)~\cite{li2019UIEB} & 3.093 & 65.664 & 3.151 & 67.927 \\
FUnIE-GAN (RA-L 2020)~\cite{islam2020funiegan} & 3.145 & 64.069 & 3.297 & 64.782 \\
USUIR (AAAI 2022)~\cite{fu2022USUIR} & 3.129 & 64.955 & 3.231 & 67.231 \\
MMLE (TIP 2022)~\cite{zhang2022MMLE} & 2.019 & 67.550 & 2.229 & 69.711 \\
Semi-UIR (CVPR 2023)~\cite{huang2023semiUIR} & 2.850 & 66.720 & 2.961 & 68.844 \\
DeepWaveNet (TOMM 2023)~\cite{sharma2021deepwavenet} & 2.706 & 66.092 & 2.722 & 68.096 \\
Histoformer (JOE 2024)~\cite{peng2024histoformer} & 3.024 & 65.278 & 3.026 & 67.918 \\
Osmosis (ECCV 2024)~\cite{nathan2024osmosis} & 2.821 & 63.294 & 2.997 & 64.089 \\
Phaseformer (WACV 2025)~\cite{khan2024phaseformer} & 2.200 & 66.810 & 2.319 & 68.824 \\
\textbf{Ours} & \textbf{3.152} & \textbf{70.110} & \textbf{3.340} & \textbf{72.457} \\
\bottomrule
\end{tabular}}
\label{tab:uiqm_musiq}
\end{table}

\begin{table}[t]
    \caption{\textbf{Qualitative evaluation on synthetic underwater dataset from~\cite{nathan2024osmosis}.} We benchmark the clear scene restoration quality on synthetic underwater images curated by~\cite{nathan2024osmosis}, which uses ground truth RGBD data from~\cite{SilbermanECCV12_nyuv2} to simulate underwater images. Our method achieves the best performance across all metrics.}
    \centering
    \resizebox{\linewidth}{!}{
    \begin{tabular}{lcccc}
    \toprule
         & PSNR $\uparrow$  & SSIM $\uparrow$ &  LPIPS $\downarrow$ & \\
    \midrule
    WaterNet (TIP 2019)~\cite{li2019UIEB}  & 18.04 & 0.75 & 0.11 &\\
    FUnIE-GAN (RA-L 2020)~\cite{islam2020funiegan} & 17.64 & 0.77 & 0.21&\\ 
    USUIR (AAAI 2022)~\cite{fu2022USUIR} & 16.76 & 0.80 & 0.18 &\\
    Semi-UIR (CVPR 2023)~\cite{huang2023semiUIR} & 17.82 & 0.83 & 0.12& \\
    MMLE (TIP 2022)~\cite{zhang2022MMLE} & 17.00 & 0.74 & 0.17 &\\
    DeepWaveNet (TOMM 2023)~\cite{sharma2021deepwavenet}  & 17.14 & 0.88 & 0.18 &\\
    Histoformer (JOE 2024)~\cite{peng2024histoformer} & 16.15 & 0.82 & 0.28 &\\
    Osmosis (ECCV 2024)~\cite{nathan2024osmosis} & 22.74 & 0.89 & 0.06 &\\
    Phaseformer (WACV 2025)~\cite{peng2024histoformer} & 16.19 & 0.78 & 0.26 &\\
    \midrule

    Ours (latent loss following~\cite{ke2023marigold, wang2024flash}) & {24.62} & {0.93} & {0.05} &\\
    Ours (w/o cross-latent decoder) & {24.99} & {0.92} & {0.06} &\\
    Ours (w/o cross-attention) & {25.06} & {0.93} & {0.05} &\\

    \textit{Ours} & \textbf{25.66} & \textbf{0.95} & \textbf{0.05} &\\
      \bottomrule
      
    \end{tabular}
    }    
    \label{tab:nyu}
\end{table}

\begin{figure}[t]
  \centering
   \includegraphics[width=\columnwidth]{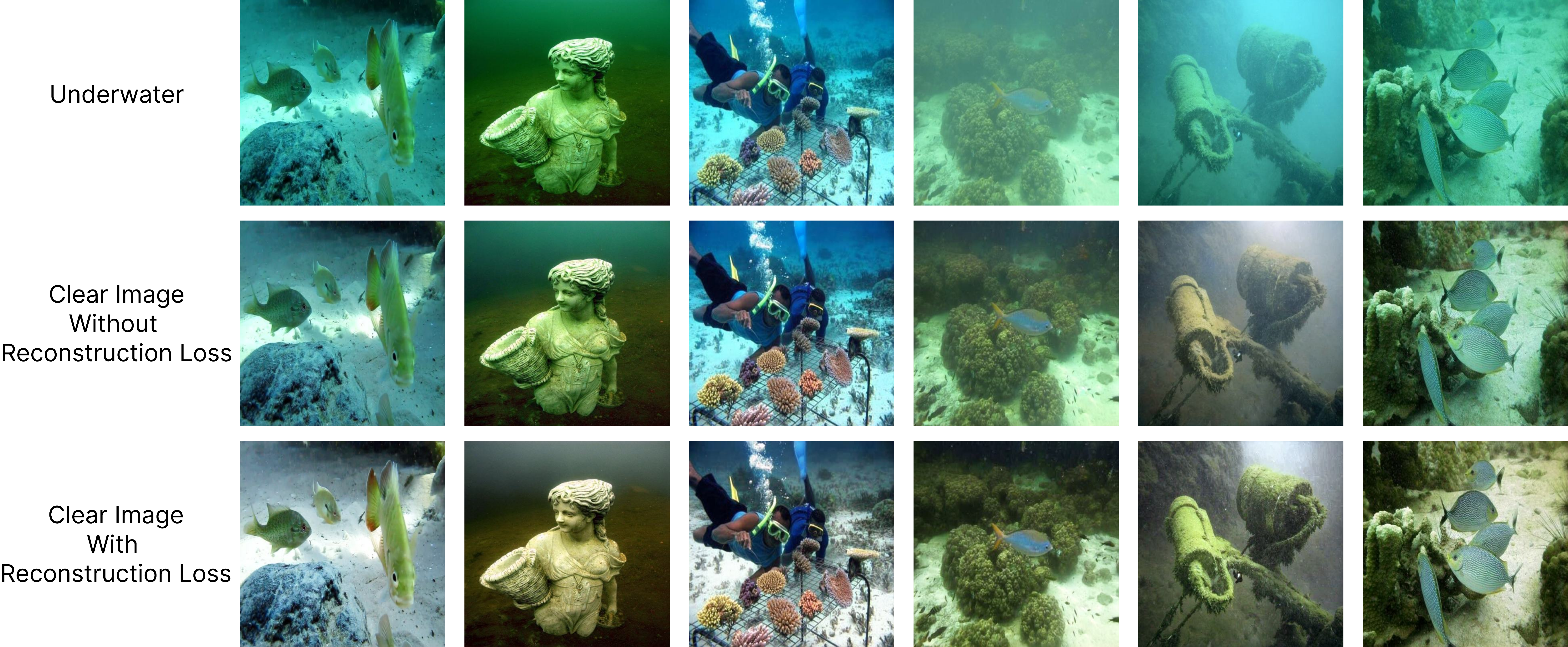}
   \caption{\textbf{Reconstruction training objective improves restoration clarity.} Our single-step fine-tuning approach enables direct end-to-end supervision on images, aligning predicted clear and medium outputs with the dense scene-medium decomposition in \cref{eq:formulation}.  Real-world evaluation on~\cite{li2019UIEB} shows that this reconstruction loss leads to clearer restorations(bottom row), compared to outputs of the model that does not use reconstruction loss (middle row).}

   \label{fig:reconstruction}
\end{figure}
\subsection{Datasets and Experiment Setups}
We use Stable Diffusion V2 (SDV2)~\cite{rombach2021latentdiffusion} diffusion UNet to initialize the scene branch, and Marigold~\cite{ke2023marigold} monocular depth model to initialize the medium branch.
We initialize the VAE weights from SDV2 pretrained weights~\cite{rombach2021latentdiffusion}.
For training data synthesis, we use high quality clean images from various sources, including natural images~\cite{Agustsson_2017DIV2K, young2014Flikr}, outdoor~\cite{zhou2017ADE20K}, indoor~\cite{zhu2022interiorverse}, night~\cite{SDV20darkzurich} images. We provide more details on training data in the Supplementary.
We train the stage 1 single-step fine-tuning and stage 2 cross-latent decoder sequentially on a single NVIDIA A6000 GPU using the same learning rate of $10^{-5}$ and $512\times 512$ image resolution. Stage 1 training took approximately 2 days and stage 2 took 1 day.

\subsection{Real World and Synthetic Comparisons}
We performed evaluations on both real-world and synthetic datasets to compare our method with existing approaches. A primary challenge in real-world data evaluation lies in the absence of ground truth clean images in available underwater datasets.
We evaluated the quality of the restored image in two reference-free metrics following previous works: UIQM~\cite{panetta2015UIQM} is tailored for underwater image restoration and measures the colorfulness, sharpness, and contrast of the restored image; MUSIQ~\cite{ke2021musiq} is a multi-scale image quality assessment metric with a transformer-based architecture.
We evaluated our method on two established real-world datasets in the underwater restoration literature: USOD10K~\cite{hong2023usod10k} and UIEB~\cite{li2019UIEB} datasets.
For a fair comparison on reference-free image quality metrics, we filtered out $\sim 10\%$ of images with apparent image artifacts unrelated to underwater effect, such as visible compression and pixelation artifacts.
Our method achieves state-of-the-art performance across all metrics on both datasets, demonstrating its effectiveness in restoring degraded underwater images to high-quality clear images.  
Our extensive qualitative comparisons in \cref{fig:qual_comp} also show that our method achieves a more distinct separation between the underlying scene content and the water medium, whereas other methods often fail to completely remove the effects of water medium, such as backscattering, or estimate them incorrectly. We provide further examples and analysis of real-world underwater images in the Supplementary.

We additionally evaluated our method on synthetic benchmarks with ground truth clear images. In~\cref{tab:nyu} we compared quantitative image metrics with baseline methods using the simulated underwater dataset in~\cite{nathan2024osmosis}, achieving the best result across all evaluated metrics. We note that none of the images in this dataset is used in our training data. These results highlight the high color accuracy and structural fidelity of our predicted clear images. 

We conducted an in-depth comparison with Osmosis~\cite{nathan2024osmosis}, the previous state-of-the-art diffusion-based underwater reconstruction method.
We show qualitative comparisons on real images in~\cite{li2019UIEB} in~\cref{fig:osmosis_cmop}.
Osmosis directly predicts depth and during iterative sampling enforces underwater image formation in~\cref{eq:UIFM_Eq}. 
However, this method is vulnerable to incorrect depth predictions, which results in spurious color patches and red shifts in the restored images that are unrealistic. 
We observe that our water-medium predictions correctly capture scene depth relations even more than the direct depth predictions of~\cite{nathan2024osmosis}, which also leads to more realistic clear image predictions.
In terms of runtime, we ran both methods using the same A6000 GPU. Due to its RGBD diffusion prior and sampling scheme, Osmosis can only restore images up to $256\times256$ size, while taking more than $200$ seconds to generate one image. In contrast, our method can restore up to $2K \times 2K$ images. Our single-step inference restores a $512 \times 512$ image in $0.75$ seconds, marking a $> 200 \times$ improvement over Osmosis.
We provide further comparisons with Osmosis in the Supplementary, including an ablation on the training dataset, and quantitative comparison on water medium prediction.

\subsection{Ablation Studies}
\label{sec:abl}

\noindent
\textbf{Single Step Prediction and Training.}
We tailor our framework to train dual-branch diffusion for single-step latent inference.
For comparison, we also train an iterative diffusion model following the fine-tuning protocol of~\cite{ke2023marigold, wang2024flash} where the diffusion UNets are supervised with a latent noise loss.
For this model, we also include cross-latent decoder training.
Synthetic results in \cref{tab:nyu} show that our single-step model performs better than the latent loss model with 50 inference steps.
This improvement stems from our ability to directly supervise the RGB output in single-step training, rather than on uninterpretable latents. 
This includes the use of reconstruction loss that enforces the clear scene and medium output to respect the dense scene-medium decomposition formulation in \cref{eq:formulation}. We show real-world examples in~\cref{fig:reconstruction} that our training objective has better scene-medium separation and produces clearer restorations.
We believe that this shows that the reconstruction loss improves our model's generalization to real-world underwater effects.

\begin{figure}[t]
  \centering
   \includegraphics[width=\columnwidth]{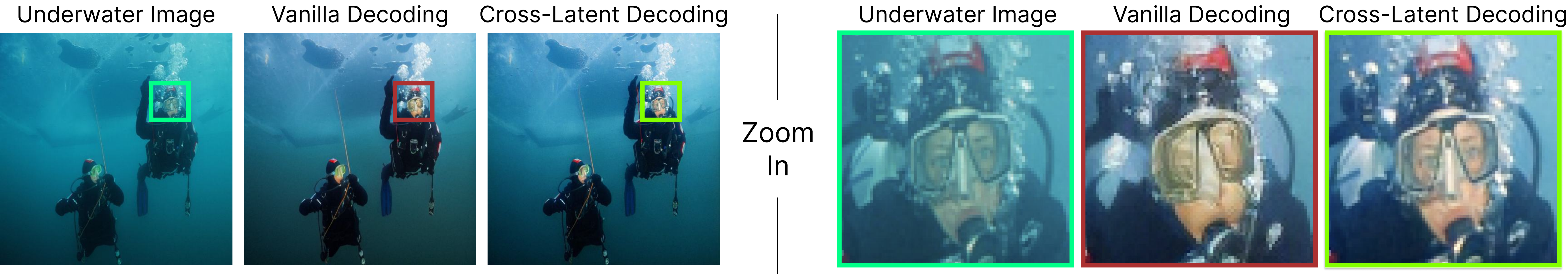}
   \caption{\textbf{Cross-latent decoding enhances restoration details.} Even though our method uses image loss during training, the limitations of the vanilla SD~\cite{rombach2021latentdiffusion} decoder could still hallucinate high-frequency details. The cross-latent decoding allows us to obtain restorations with better details, such as the eyes of the diver in this figure.}

   \label{fig:cross_decoder}
\end{figure}
\vspace{1mm}
\noindent
\textbf{Cross-Latent Decoder.} 
Even when guided by an image reconstruction loss during training, the standard SD decoder~\cite{rombach2021latentdiffusion} can sometimes introduce hallucinations in detailed regions due to the inherent challenges of reconstructing detail from a compressed latent space. 
As shown in~\cref{fig:cross_decoder}, the enhanced detail in the diver's eyes illustrates the effectiveness of cross-latent decoder in preserving critical high-frequency information, particularly in regions where the vanilla SD decoder might struggle.

\section{Limitations}
While our method effectively restores high-quality images and estimates water parameters from single underwater inputs, it has a few limitations. Although faster and more efficient than prior diffusion-based approaches, it still requires a consumer-grade GPU and does not yet achieve real-time performance. Additionally, as it operates on single images, temporal consistency is not enforced, which we demonstrate on the MKV underwater video dataset~\cite{MVK} in the Supplementary.

\section{Conclusion}
In conclusion, we propose a novel underwater image restoration framework that leverages the foundational natural image and geometric priors embedded in pretrained latent diffusion models. Our approach introduces a fast, single-step restoration pipeline capable of producing detailed and robust predictions of both the clear scene and the intervening water medium. 
To train our model, we develop a physically grounded underwater image synthesis pipeline that generates realistic and diverse synthetic training data at scale. Comprehensive experiments on both synthetic and real-world benchmarks demonstrate that our method achieves state-of-the-art restoration performance, significantly advancing the quality and efficiency of diffusion-based underwater image restoration.

\ifpeerreview \else
\section*{Acknowledgments}
J.W. and Y.A. were supported in part by USDA NIFA sustainable agriculture system program under award no.~20206801231805.
T.W. and C.A.M. were supported in part by the UMD AIM Seed Grant Program, NSF CAREER grant no.~2339616, and ONR grant no.~N00014-23-1-2752. 
M.A.S. and M.J.I. were supported in part by the NSF grant no.~2330416.
\fi

\bibliographystyle{IEEEtran}
\bibliography{references}

\ifpeerreview \else

\begin{IEEEbiography}
[{\includegraphics[width=1in,height=1.25in,clip,keepaspectratio]{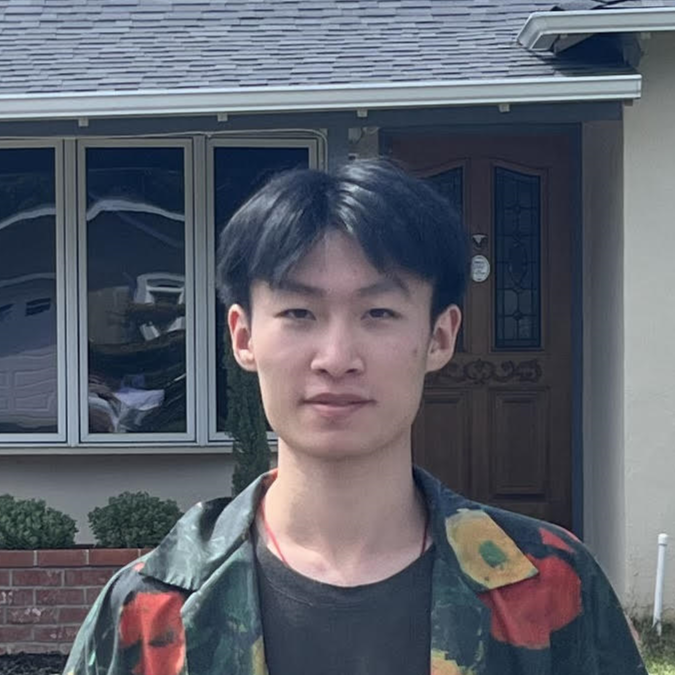}}]{Jiayi Wu} is a Ph.D. student at the Perception and Robotics Group of the University of Maryland, College Park. He received his M.Sc. (2023) in ECE from the University of Florida. His research focuses on 3D/4D generation, differentiable rendering, and active vision.
\end{IEEEbiography}

\begin{IEEEbiography}[{\includegraphics[width=1in,height=1.25in,clip,keepaspectratio]{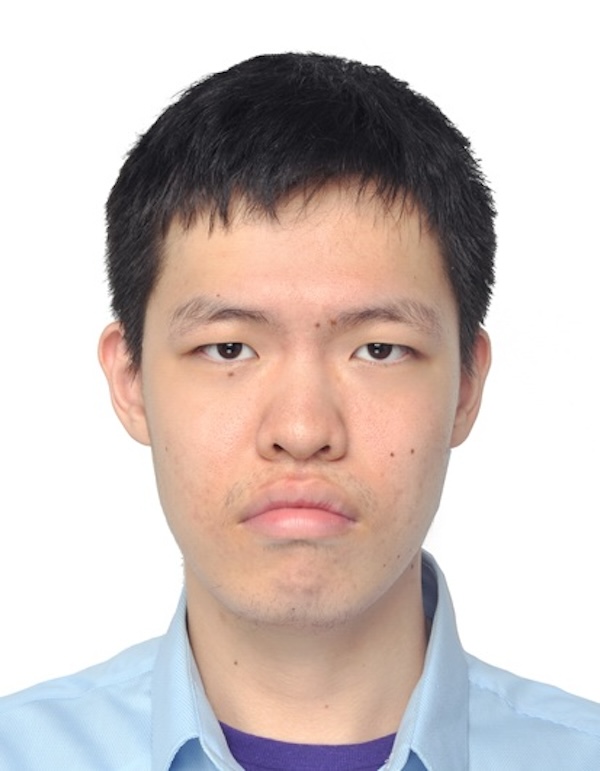}}]{Tianfu Wang}
is a Ph.D. student at the Intelligent Sensing Lab of the University of Maryland, College Park. Tianfu completed his Master's degree in Computer Science at ETH Zurich, and his Bachelor's degree in Computer Science at Northwestern University. Tianfu is interested in computational imaging, generative models, and differentiable rendering.  
\end{IEEEbiography}

\begin{IEEEbiography}[{\includegraphics[width=1in,height=1.25in,clip,keepaspectratio]{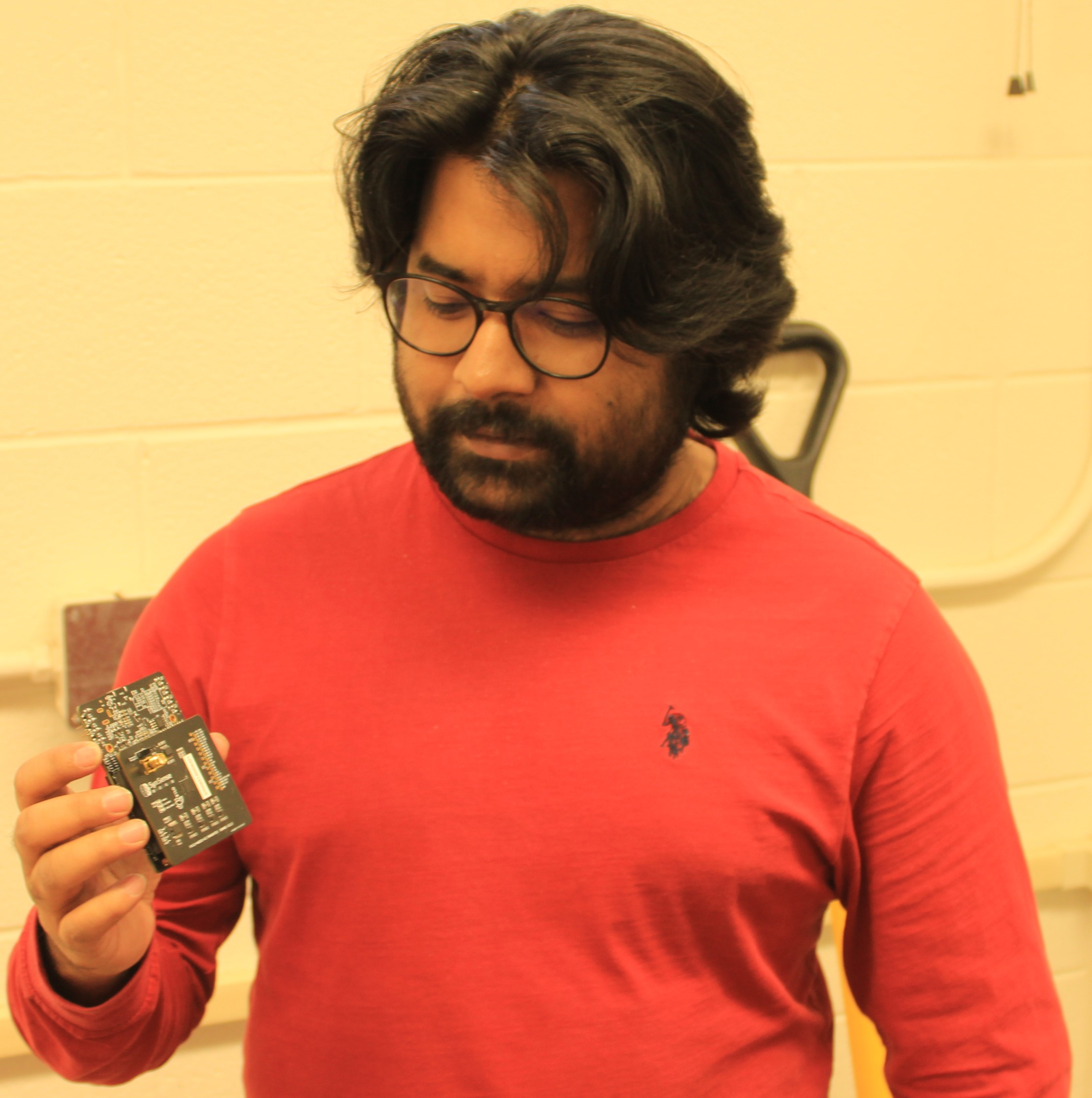}}]{Md Abu Bakr Siddique} is a Ph.D. student at the dept of ECE, University of Florida. He completed an M.Sc. in ECE from Michigan Technological University (2024), and B.Sc. in EEE from IUT, Bangladesh. Post graduation, Abu worked as a faculty member at IUBAT, Bangladesh. His research interest lies in the intersection of Machine Learning and Robotics. He is exploring 3D reconstruction and mapping of underwater scenes. He is also exploring the visual servoing and image-guided exploration by autonomous mobile robots.  
\end{IEEEbiography}

\begin{IEEEbiography}[{\includegraphics[width=1in,height=1.25in,clip,keepaspectratio]{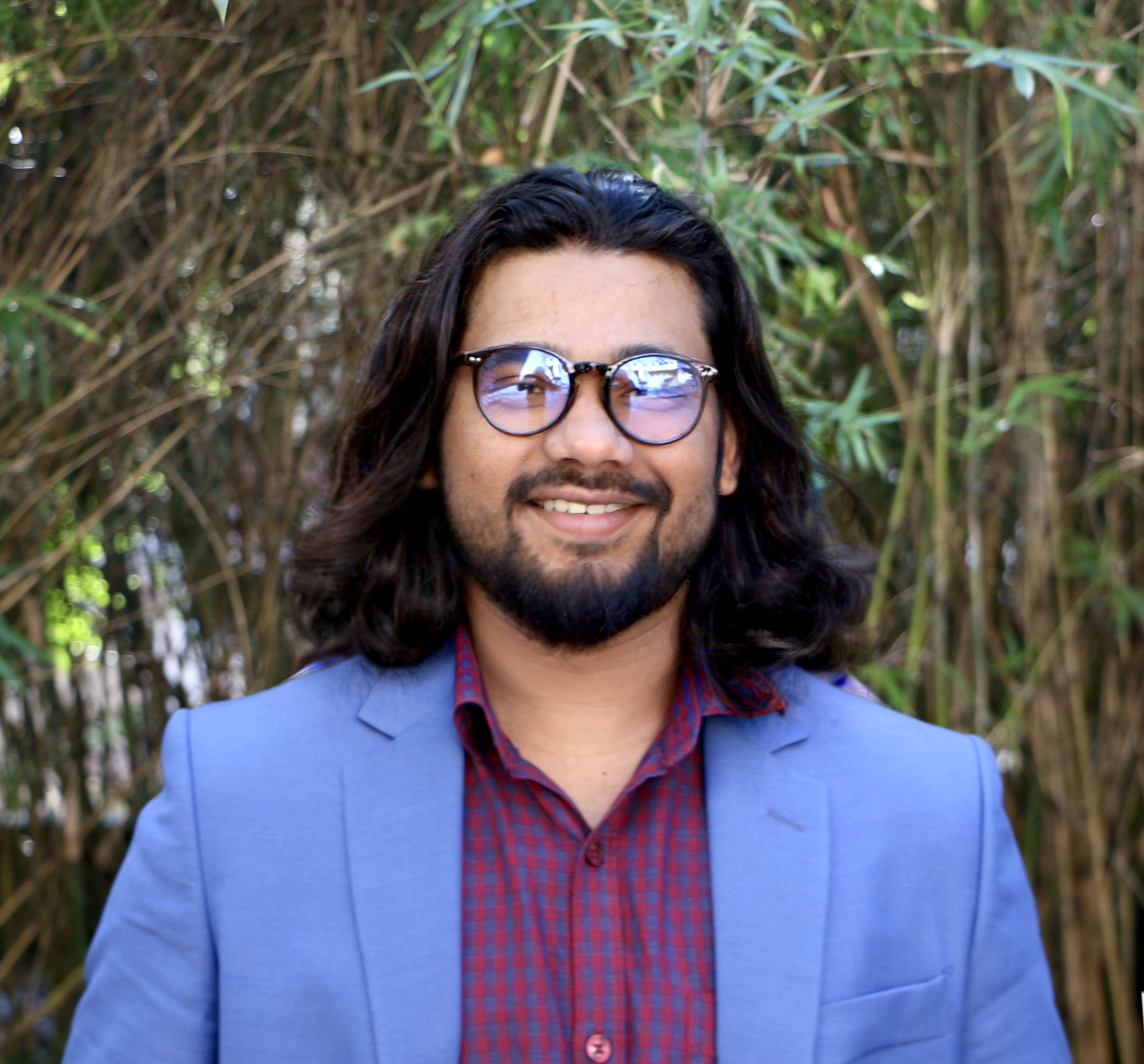}}]{Md Jahidul Islam} is an Assistant Professor at the Department of ECE of the University of Florida (UF). He received his Ph.D. (2021) in Robotics from the University of Minnesota. His research focuses on enabling active perception and navigation of autonomous underwater robots. He leads the RoboPI group toward developing next-generation robotics systems for subsea inspection, surveillance, and monitoring; his current projects are funded by multiple projects from the NSF, ONR, and TI.
\end{IEEEbiography}

\vspace{50pt}

\begin{IEEEbiography}[{\includegraphics[width=1in,height=1.25in,clip,keepaspectratio]{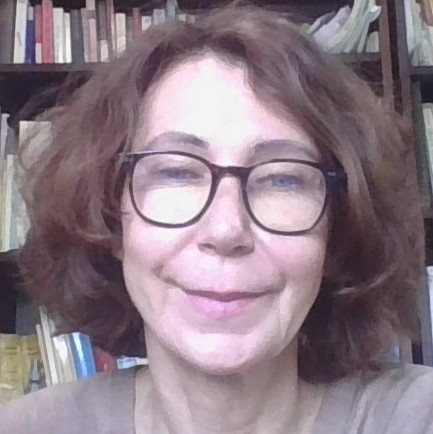}}]{Cornelia Fermüller} is a research scientist at the Institute for Advanced Computer Studies (UMIACS) at the University of Maryland at College Park.  She holds a Ph.D. from the Technical University of Vienna, Austria, and an M.S. from the University of Technology, Graz, Austria, both in Applied Mathematics.  Her research is in Computer, Human, and Robot Vision. She studies and develops biologically inspired Computer Vision solutions for systems that interact with their environment. In recent years, her work has focused on interpreting human activities in the context of music education and on motion processing for fast active robots using bio-inspired event-based sensors as input.
\end{IEEEbiography}

\begin{IEEEbiography}[{\includegraphics[width=1in,height=1.25in,clip,keepaspectratio]{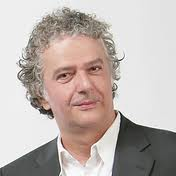}}]{Yiannis Aloimonos} is Professor of Computational Vision and Intelligence at the Department of Computer Science, University of Maryland, College Park, and the Director of the Computer Vision Laboratory at the Institute for Advanced Computer Studies (UMIACS). He is also affiliated with the Institute for Systems Research and the Neural and Cognitive Science Program. He was born in Sparta, Greece and studied Mathematics in Athens and Computer Science at the University of Rochester, NY (PhD 1990). He is interested in Active Perception and the modeling of vision as an active, dynamic process for real time robotic systems. For the past five years he has been working on bridging signals and symbols, specifically on the relationship of vision to reasoning, action and language. He received the Presidential Young Investigator Award from President G. Bush. He is an IEEE Fellow.
\end{IEEEbiography}

\begin{IEEEbiography}[{\includegraphics[width=1in,height=1.25in,clip,keepaspectratio]{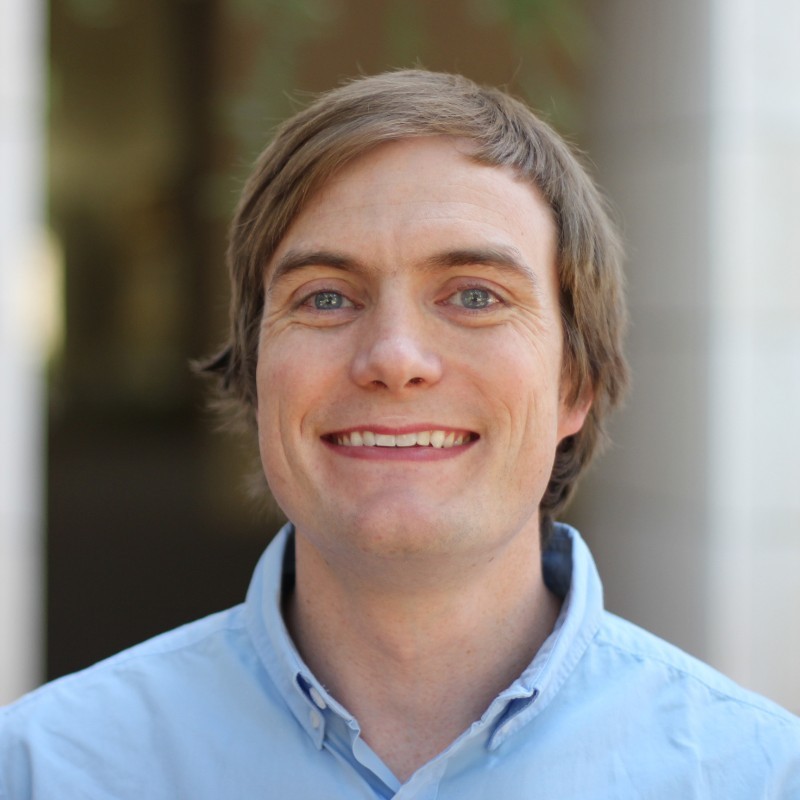}}]{Christopher A. Metzler} is an Assistant Professor in the Department of Computer Science at
the University of Maryland College Park, where he leads the UMD Intelligent Sensing Laboratory. He is a member of the University of Maryland Institute for Advanced Computer Studies (UMIACS) and has a courtesy appointment in the Electrical and Computer Engineering Department. His research develops new systems and algorithms for solving problems in computational imaging and sensing, machine learning, and wireless communications. His work has received multiple best paper awards; he recently received NSF CAREER, AFOSR Young Investigator Program, and ARO Early Career Program awards; and he was an Intelligence Community Postdoctoral Research Fellow, an NSF Graduate Research Fellow, a DoD NDSEG Fellow, and a NASA Texas Space
Grant Consortium Fellow.
\end{IEEEbiography}

\fi

\clearpage

\clearpage
\twocolumn[{
  \centering
  \normalfont\normalsize\vskip0.2em{\Huge Supplementary Material for: ``Single-Step Latent Diffusion for Underwater Image Restoration" \par}\vskip1.0em\par
}]

\begin{figure}[t]
  \centering
   \includegraphics[width=\columnwidth]{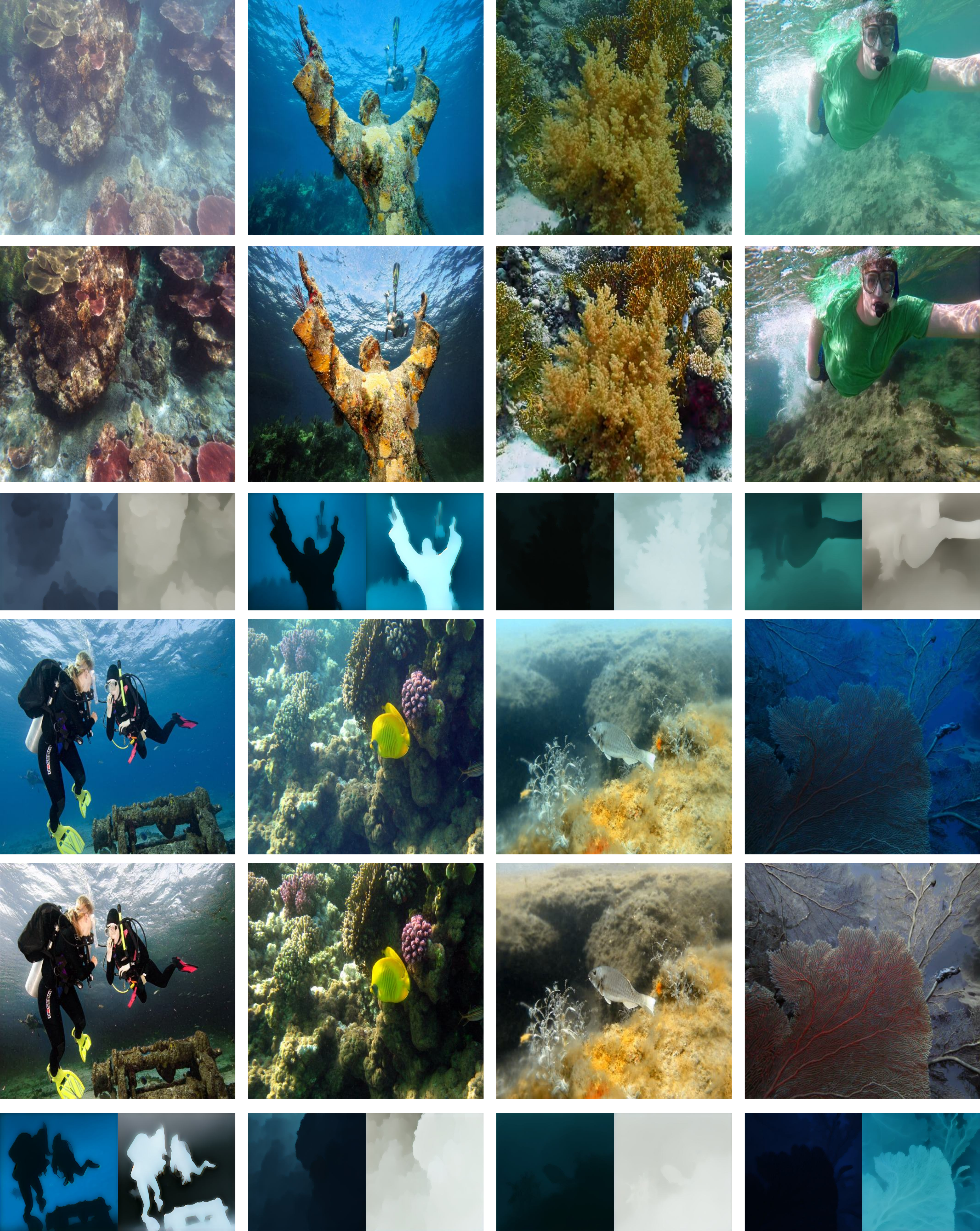}
   \vspace{-20pt}
   \caption{\textbf{More restoration results of our method on real-world datasets~\cite{hong2023usod10k, li2019UIEB}.} We showcase more real-world restoration image results of our method that we find visually appealing.  }

   \label{fig:supp_showcase}
\end{figure}

\section{Comparison with Latent Loss Training Methods}
\begin{table}[t]
\vspace{-5pt}

    \centering
    \caption{\textbf{Comparing our single-step method to iterative latent loss fine-tuning methods.} We present quantitative comparisons on the synthetic dataset of~\cite{nathan2024osmosis} for our single-step diffusion restoration method, and iterative diffusion (50 steps) diffusion method fine-tuned following the latent loss of Marigold~\cite{ke2023marigold, wang2024flash}. Our results demonstrate that our single-step restoration model outperforms latent-loss-based models using 50 denoising steps, showing that the effectiveness of our training pipeline goes beyond merely accelerating inference speed.
}
    \resizebox{\linewidth}{!}{
    \begin{tabular}{lcccc}
    \toprule
       & PSNR $\uparrow$  & SSIM $\uparrow$ &  LPIPS $\downarrow$ & \\
    \midrule
    Iterative Latent Loss (50 Steps) & {23.46} & {0.89} & {0.08} &\\
    Iterative Latent Loss (50 Steps) + CLD & {24.62} & {0.93} & {0.05} &\\
    \textit{Ours} (Single-Step) & \textbf{25.66} & \textbf{0.95} & \textbf{0.05} &\\
      \bottomrule
      
    \end{tabular}
    }    
    \vspace{-5pt}

    \label{tab:sup_lat}
\end{table}

Our single-step pipeline not only provides faster inference speeds, but also enables the use of image losses during training. Previous fine tuning methods for multi-step iterative inference, such as Marigold~\cite{ke2023marigold, wang2024flash}, use a latent space loss, which is less interpretable. To show the advantage of our one-step architecture and training objectice, we present quantitative results for restoration models trained using the standard diffusion latent loss objective with 50 denoising steps used in Marigold~\cite{ke2023marigold, wang2024flash}, with and without the cross-latent decoder (CLD). Our results in~\cref{tab:sup_lat} show that our single-step model outperforms multi-step diffusion fine-tuned with  latent loss.   
Finally, we highlight several key limitations regarding few-step diffusion distillation methods~\cite{salimansprogressive,song2023consistency, luo2023latent}.
Distillation requires training an iterative diffusion teacher model before distilling it into a single-step model, while our method is trained for single-step prediction from the outset.
Moreover, previous work~\cite{salimansprogressive,song2023consistency, luo2023latent} has consistently shown that the output quality of single-step distilled models is upper bounded by their iterative teacher models. In contrast, our single-step model already outperforms the multi-step diffusion.

\section{Accuracy of Medium Prediction}
Due to the lack of ground truth data on transmission and backscattering, we follow Osmosis~\cite{nathan2024osmosis} and evaluate our method on the unseen simulated NYU dataset~\cite{SilbermanECCV12_nyuv2}. In~\cref{tab:supp_tb}, we measure the PSNR and the MAE of the predicted transmission (T) and backscattering (B) compared to the ground truth. We achieve superior medium prediction accuracy for both predictions over Osmosis~\cite{nathan2024osmosis}, which struggles due to its unreliable depth prediction similar to~\cref{fig:osmosis_cmop} in the main paper.

\begin{table}[t]
\vspace{-5pt}
    \centering
    \caption{\textbf{Quantitative evaluation on water medium prediction.}
     We report PSNR and MAE for predicted transmission (T) and backscattering (B) against the ground truth on the simulated NYU~\cite{SilbermanECCV12_nyuv2} underwater dataset. Our method achieves higher accuracy for both components compared to  Osmosis~\cite{nathan2024osmosis}, which suffers from unreliable depth estimation.}
    \resizebox{\linewidth}{!}{
    \begin{tabular}{lccccc}
    \toprule
     & PSNR T. $\uparrow$  & MAE T. $\downarrow$  &  PSNR B. $\uparrow$ & MAE B. $\downarrow$ \\
    \midrule
    Osmosis & {13.97} & {0.207} & {23.08} &{0.076}\\
    Ours  & \textbf{24.69} & \textbf{0.060} & \textbf{32.37} & \textbf{0.024}\\
      \bottomrule
      
    \end{tabular}
    }    
    \label{tab:supp_tb}
    \vspace{-5pt}

\end{table}

\section{Additional Ablation Quantitative Comparisons}
We report quantitative results on the simulated dataset from Osmosis~\cite{nathan2024osmosis, SilbermanECCV12_nyuv2} for our ablation studies presented in \cref{sec:abl} of the main paper in~\cref{tab:supp_abl}, specifically the use of reconstruction loss enabled by our single-step training, and the cross-latent decoder (CLD). Reconstruction loss improves the PSNR by 0.86 dB and the cross-latent decoder improves the PSNR by 0.67 dB. 
\begin{table}[t]
    \vspace{-5pt}  
    \centering
    \caption{\textbf{Quantitative results for ablation studies in the main paper.} We provide further quantitative results on the synthetic dataset of~\cite{nathan2024osmosis} for the ablation studies in \cref{sec:abl} of the main paper.}
    \resizebox{\linewidth}{!}{
    \begin{tabular}{lcccc}
    \toprule
     & PSNR $\uparrow$  & SSIM $\uparrow$ &  LPIPS $\downarrow$ & \\
    \midrule
    Ours without Recon. Loss & {24.80} & {0.93} & {0.05} &\\
    Ours without CLD & {24.99} & {0.92} & {0.06} &\\
    {Ours} & \textbf{25.66} & \textbf{0.95} & \textbf{0.05} &\\
      \bottomrule
    \end{tabular}
    }    
    \vspace{-5pt}      

    \label{tab:supp_abl}
\end{table}

\begin{table*}[h]
\vspace{-5pt}
    \centering
    \caption{\textbf{Ablation study on the effect of model architecture and data.} We decouple our physics-based diverse underwater data synthesis pipeline and our single-step restoration network to further study how each component affects the performance of our method. Specifically, we train our method using the terrestrial dataset from Osmosis~\cite{nathan2024osmosis} instead of our terrestrial data, and we use randomized water medium parameters instead of our curated values from real-world measurements. We show quantitative results on the simulated underwater dataset using~\cite{SilbermanECCV12_nyuv2} from~\cite{nathan2024osmosis}. Our results show that using real-world water medium values during training data synthesis boosts the restoration accuracy of our model. On the other hand, even with randomized water parameters and using the same training data as Osmosis~\cite{nathan2024osmosis}, our method still outperforms the baseline.}
    \resizebox{0.9\linewidth}{!}{
    \begin{tabular}{lccccc}
    \toprule
     Method & Terrestrial Data & Water Medium Data & PSNR  $\uparrow$  & SSIM $\uparrow$  & LPIPS  $\downarrow$   \\
    \midrule
    Osmosis & Osmosis Data & - &{22.74} & {0.89} & {0.06}\\
    Ours  & Osmosis Data & Random Values &{24.87} & {0.94} & {0.05} \\
    Ours  & Osmosis Data & Sample From Real-World Measurements & \textbf{25.76} & {0.95} & {0.05} \\
    Ours  & Our Data & Sample From Real-World Measurements &{25.66} & \textbf{0.95} & \textbf{0.05} \\
      \bottomrule
      
    \end{tabular}
    }    
    \label{tab:supp_data}
    \vspace{-5pt}

\end{table*}

\section{Disentangling the Impact of Network Design and Training Data}

Our work consists of two key components: a novel single-step diffusion underwater restoration network, and a physics-based diverse underwater training data synthesis pipeline. To further investigate the effect for each component on our method's final performance, we conduct an additional ablation study that disentangles data and network. 
Specifically, we train new versions of our network using the same RGBD terrestrial data that Osmosis~\cite{nathan2024osmosis} used for its RGBD diffusion prior.
We also set the water parameters \( \beta^D \), \( \beta^B \), and \( B^\infty \) to random RGB values, instead of sampling from real-world water measurements. 
Our results in~\cref{tab:supp_data} show that using real-world water parameters boosts the performance of the trained restoration model, showing the strength to integrate domain-specific water medium knowledge to training. 
However, even after training with random water medium parameters, our method still outperforms Osmosis and other baselines (see \cref{tab:nyu} of the main paper), demonstrating the effectiveness of our single-step diffusion network and the underlying diffusion priors for underwater image restoration.

\section{Robustness to Challenging Underwater Scenarios}
\begin{figure}[h]
  \centering
  \includegraphics[width=\columnwidth]{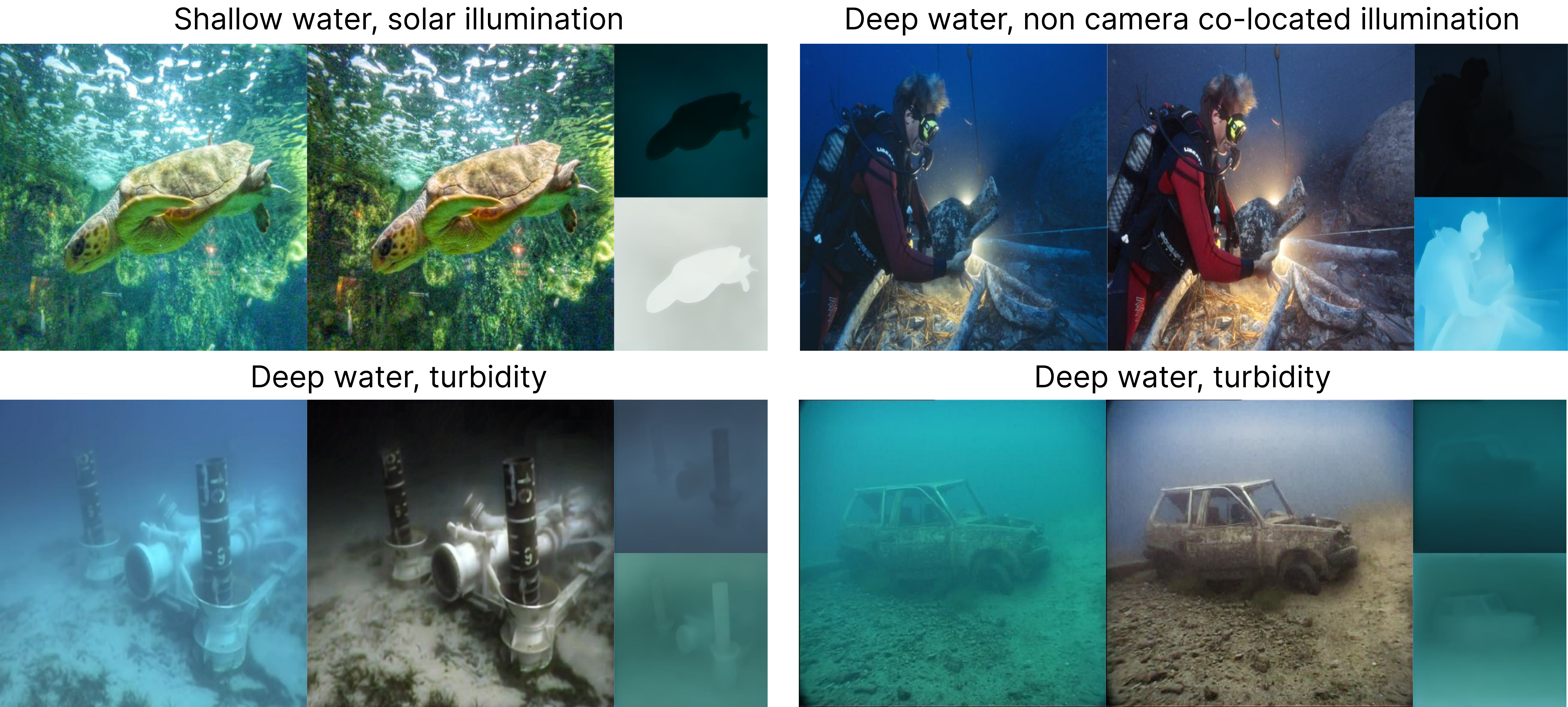}

  \caption{\textbf{Visualizing our model's performance under challenging underwater scenarios.} We show method restoring real world examples from~\cite{li2019UIEB, hong2023usod10k} with challenging lighting scenarios and strong turbidity. }
\label{fig:supp_challenging}
   
\end{figure}

Our simulation pipeline is based on the underwater image formation model in~\cref{eq:UIFM_Eq}, which models scattering of light from the object surface without assuming co-location of the illumination source and the sensor. While~\cref{eq:UIFM_Eq} does not explicitly account for complex underwater phenomena such as turbidity, our prediction framework extends this formulation by leveraging the more flexible dense scene-medium decomposition in~\cref{eq:formulation}. To evaluate robustness of our formulation and our pipeline, we show real-world examples in~\cref{fig:supp_challenging} spanning a range of conditions including shallow water under solar illumination, deep water with non co-located light sources, and scenes with noticeable turbidity. Our method demonstrates strong performance across these diverse scenarios, although blurriness may appear in cases of severe turbidity. Incorporating a more explicit modeling of turbidity into our formulation presents a promising avenue for future research.

\section{More Real-World Results}

\begin{figure*}[t]
  \centering
   \includegraphics[width=\linewidth]{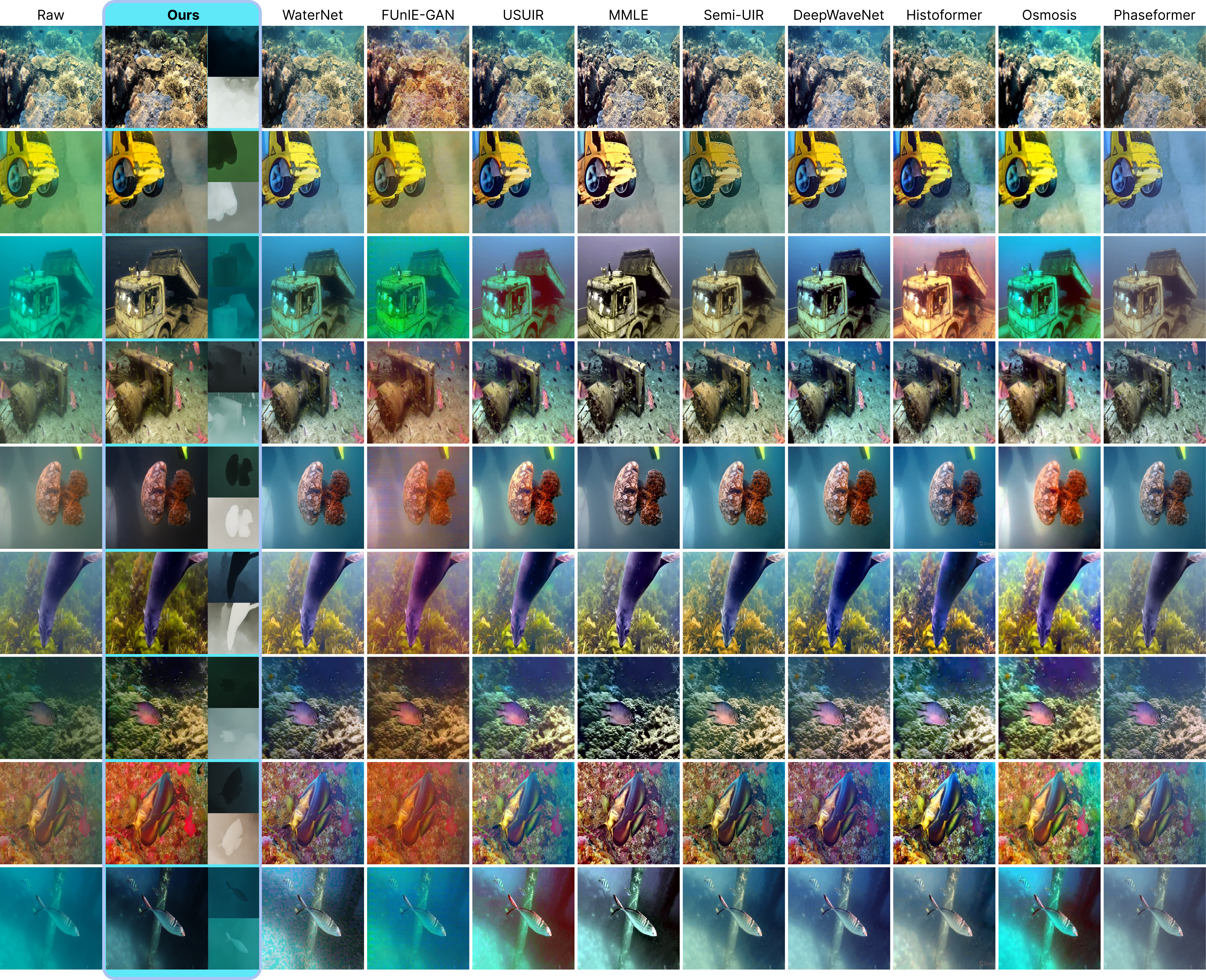}
   \vspace{-20pt}
   \caption{\textbf{Our method restores more accurate colors compared to other methods.}
When evaluating restored outputs on real-world underwater datasets~\cite{li2019UIEB, hong2023usod10k}, we observe that our method more accurately recovers the true color profiles of objects in both the foreground and background. In contrast, other methods often introduce artifacts such as spurious red shifts, overcompensation for medium effects, or incomplete removal of background light. We attribute our improved color fidelity to the strong natural image priors inherent in pretrained latent diffusion models~\cite{rombach2021latentdiffusion}, combined with our physically informed fine-tuning objectives.   }

   \label{fig:supp_color_shift}
\end{figure*}

\begin{figure*}[t]
  \centering
   \includegraphics[width=\linewidth]{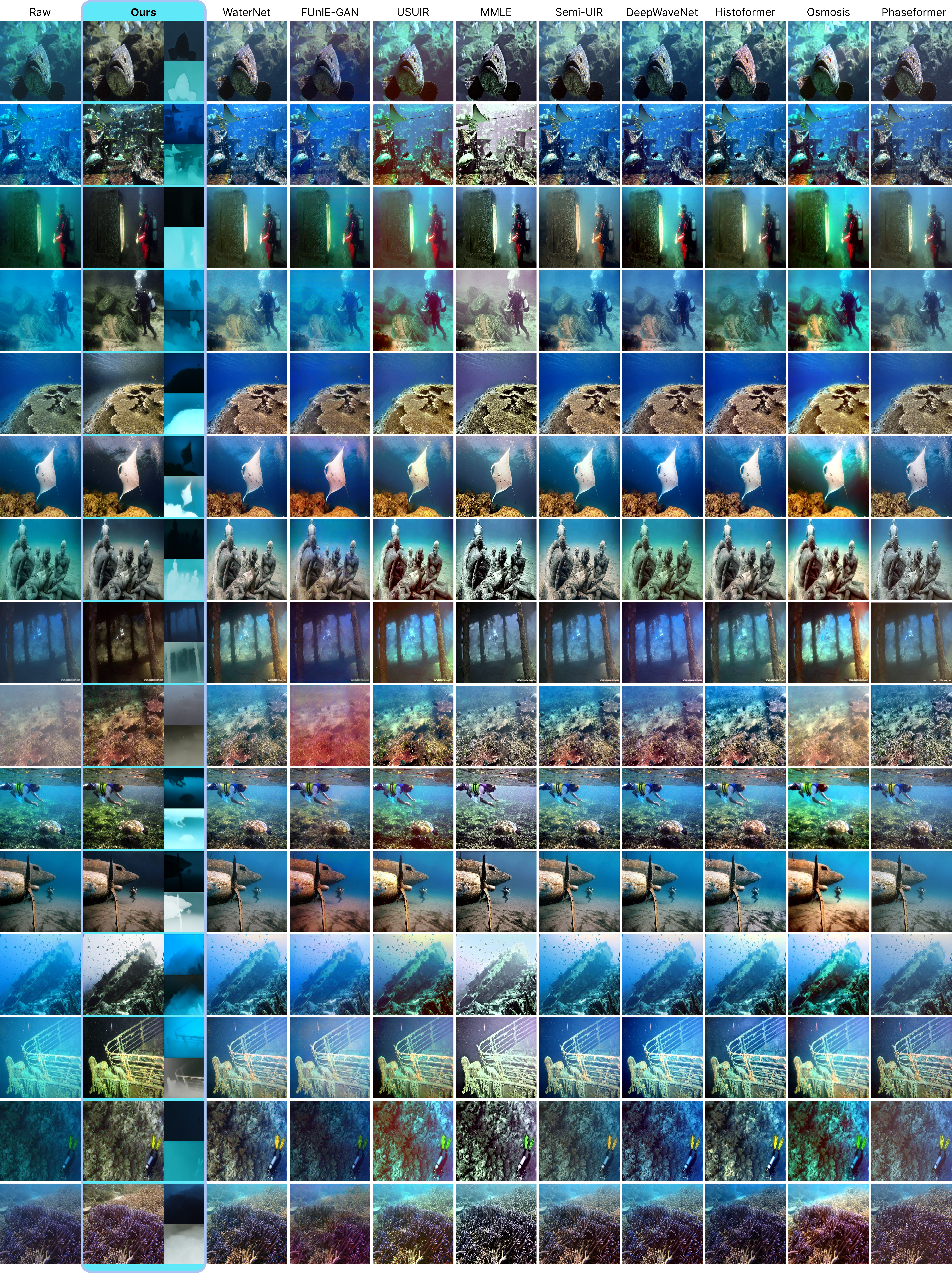}
   \vspace{-20pt}
   \caption{\textbf{Additional real-world comparisons on real-world underwater datasets~\cite{hong2023usod10k, li2019UIEB}.}  Our method (second column from the left) achieves physically consistent results across varying depths, with improved performance in degraded distant regions. It also accurately estimates per-pixel medium parameters across diverse water types, enabling faithful scene restoration under various underwater conditions. }

   \label{fig:supp_add_results}
\end{figure*}

In our qualitative comparisons on real-world underwater datasets~\cite{li2019UIEB, hong2023usod10k}, we observed a consistent trend in the color accuracy of our restored images. We show this effect in~\cref{fig:supp_color_shift} on a wide range of examples. While evaluating samples across a wide range of scenes, our method consistently recovered more faithful color profiles for both foreground and background objects. In contrast, other approaches frequently introduced color distortions, such as unnatural red shifts, overcompensation for underwater effects, or incomplete removal of background light. We attribute this advantage to the strong natural image priors embedded in the pretrained latent diffusion backbone~\cite{rombach2021latentdiffusion}, our scene-medium decomposition formulation, as well as our physically informed fine-tuning objectives, which together enable more precise modeling of underwater image degradation and restoration.
Finally, we show additional real-world scenes in~\cref{fig:supp_showcase}  and  comparison results in~\cref{fig:supp_add_results}.

\section{More Synthetic Training Data Examples}
\begin{figure*}[t]
  \centering
   \includegraphics[width=\linewidth]{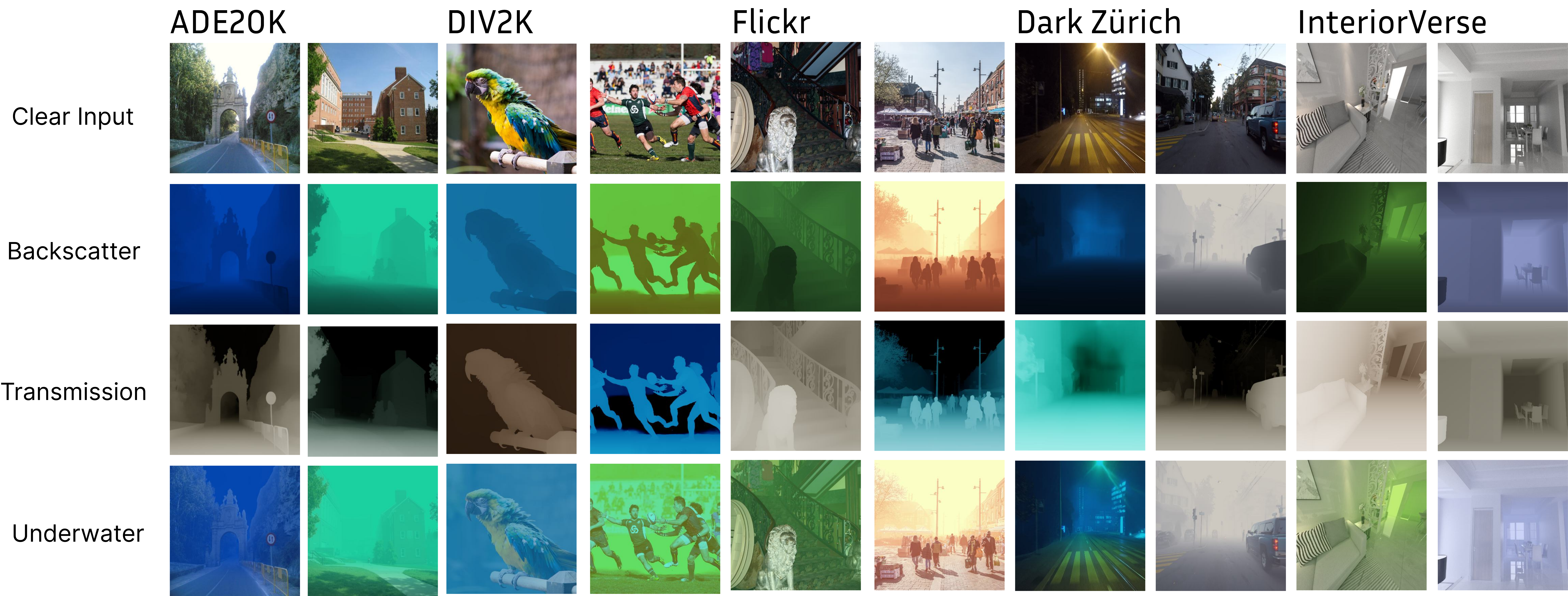}
   \vspace{-20pt}
   \caption{\textbf{More visualizations of synthetic training data.} We show extensive visualizations of training data samples. We use a wide range of terrestrial image data sources as the clean image. Combined with our  physically-accurate data synthesis pipeline, we generate diverse and realistic underwater images with various underwater medium profiles. ADE20K~\cite{zhou2017ADE20K} is an outdoor dataset originally used for semanitic segmentation. DIV2K~\cite{Agustsson_2017DIV2K} is a high-resolution dataset with diverse scene content. Flickr dataset~\cite{young2014Flikr} is a large dataset consisting of crowd-sourced Internet images. Dark Zurich~\cite{SDV20darkzurich} is a dataset of city scenes in night environment. InteriorVerse~\cite{zhu2022interiorverse} is an indoor dataset.}

   \label{fig:supp_train_data}
\end{figure*}

In~\cref{fig:supp_train_data} we present a diverse set of visualizations illustrating the synthetic underwater training data generated using our physically-accurate data synthesis pipeline detailed in~\cref{sec:data} of the main paper. To ensure diversity and realism, we source clean images from a wide range of large-scale terrestrial datasets spanning both indoor and outdoor environments. These include ADE20K~\cite{zhou2017ADE20K}, an outdoor dataset originally designed for semantic segmentation; DIV2K~\cite{Agustsson_2017DIV2K}, a high-resolution dataset containing diverse photographic scenes; and the Flickr dataset~\cite{young2014Flikr}, which comprises a broad collection of crowd-sourced Internet images. We also incorporate Dark Zurich~\cite{SDV20darkzurich}, which features urban street scenes captured in low-light, nighttime conditions, and InteriorVerse~\cite{zhu2022interiorverse}, a dataset focused on richly detailed indoor environments. 
Using our synthesis pipeline, we simulate a variety of underwater conditions by varying medium parameters such as depth, attenuation, and background light, resulting in a rich and diverse training dataset that better captures the complexity of real-world underwater imaging scenarios. The visualizations in~\cref{fig:supp_train_data} also highlight the diversity of underwater medium profiles, showcasing that our data synthesis pipeline is able to capture a wide range of water types and lighting conditions to reflect the variability found in natural underwater environments.


\section{Video Results}
While our method is designed for single-image restoration and does not explicitly model temporal consistency, we evaluate it on underwater video sequences from the MVK dataset~\cite{MVK} to assess its performance across frames. 
\textcolor{blue}{We refer readers to the static HTML file provided in the supplementary files (supp\_video.html) or directly in the "videos" folder to viewing of the video restoration results.}
The videos demonstrate restoration results in diverse underwater environments, from shallow reefs to deep-sea and wreck scenes. Our method yields stable, view-consistent outputs for foreground objects with minimal flickering, even for moving objects. However, in scenes with large depth variation or low light, flickering artifacts emerge between frames. 
A notable observation is that the model adapts well to focus changes in deep-sea footage, producing clearer predictions once objects come into focus. The failure case in the wreck video highlights limitations in our current approach, with significant flickering attributed to the absence of temporal modeling, noisy inputs, and ambiguous depth cues from particles in the scene.
We believe that the flickering artifacts in our videos are temporal in nature and stems from the fact that our method does not model temporal consistency.
Our foreground objects maintain strong view consistency across frames. On a per-frame basis, the restoration quality of the overall image is high with no apparent artifacts.
Overall, these videos demonstrate the method’s strong performance in restoring underwater visuals from single frames, while also motivating future work to incorporate temporal consistency for video-based applications.


\end{document}